\documentclass{article}
\usepackage[square,numbers,sort]{natbib}
\usepackage[left=1in,right=1in,top=1in,bottom=1in]{geometry}

\usepackage[utf8]{inputenc} 
\usepackage[T1]{fontenc}    
\usepackage[colorlinks=true,citecolor=blue,urlcolor=black]{hyperref}       
\usepackage{url}            
\usepackage{booktabs}       
\usepackage{amsfonts}       
\usepackage{nicefrac}       
\usepackage{microtype}      
\usepackage[dvipsnames]{xcolor}         

\usepackage{xspace} 
\usepackage{enumitem} 
\usepackage{multirow}
\usepackage{graphicx}
\usepackage[labelfont=bf,font=normalsize]{caption}
\usepackage{subcaption}
\usepackage{wrapfig}
\usepackage{cleveref}
\makeatletter
\newcommand\footnoteref[1]{\protected@xdef\@thefnmark{\ref{#1}}\@footnotemark}
\makeatother

\newcommand{\name}{{\texttt{Motley}}\xspace}

\newcommand{\codeurl}{\url{https://github.com/google-research/federated/tree/master/personalization_benchmark}}

\usepackage{bm}
\usepackage{arydshln}

\makeatletter
\def\Cline#1#2{\@Cline#1#2\@nil}
\def\@Cline#1-#2#3\@nil{%
  \omit
  \@multicnt#1%
  \advance\@multispan\m@ne
  \ifnum\@multicnt=\@ne\@firstofone{&\omit}\fi
  \@multicnt#2%
  \advance\@multicnt-#1%
  \advance\@multispan\@ne
  \leaders\hrule\@height#3\hfill
  \cr}
\makeatother

\title{\name: Benchmarking Heterogeneity and Personalization in Federated Learning}

\author{%
  Shanshan Wu$^1$, Tian Li$^2$, Zachary Charles$^1$, Yu Xiao$^1$,\\
  Ziyu Liu$^2$, Zheng Xu$^1$, Virginia Smith$^2$\\
  $^1$Google Research \texttt{\{shanshanw, zachcharles, xiaoyux, xuzheng\}@google.com} \\
  $^2$Carnegie Mellon University  \texttt{\{tianli, kzliu, smithv\}@cmu.edu}
}

\date{}

\begin{document}

\maketitle

\begin{abstract}
Personalized federated learning considers learning models unique to each client in a heterogeneous network. The resulting client-specific models have been purported to improve metrics such as accuracy, fairness, and robustness in federated networks.  However, despite a plethora of work in this area, it remains unclear: (1) which personalization techniques are most effective in various settings, and (2) how important personalization truly is for realistic federated applications. To better answer these questions, we propose \name, a benchmark for personalized federated learning. \name consists of a suite of cross-device and cross-silo federated datasets from varied problem domains, as well as thorough evaluation metrics for better understanding the possible impacts of personalization. We establish baselines on the benchmark by comparing a number of representative personalized federated learning methods. These initial results highlight strengths and weaknesses of existing approaches, and raise several open questions for the community. \name aims to provide a reproducible means with which to advance developments in personalized and heterogeneity-aware federated learning, as well as the related areas of transfer learning, meta-learning, and multi-task learning. 
\end{abstract}
\section{Introduction}\label{sec:intro}
Federated learning (FL) aims to share knowledge across disparate data silos in a privacy-preserving manner~\cite{mcmahan2017communication,lireviewpaper,kairouz2021advances}. Relative to standard (e.g., data center) distributed learning, a defining trait of federated learning is the presence of \textit{heterogeneity}\footnoteref{hetero}, as each of the clients (e.g., a sensor, mobile phone, or entire organization) may generate data according to a distinct distribution. In response to this key difference, a significant amount of work has been devoted to understanding and addressing heterogeneity in federated learning. For example, heterogeneity has been the focus of the analysis and development of numerous federated optimization methods~\cite[e.g.,][]{li2018federated,karimireddy2020scaffold,hsu2019measuring,li2019convergence,charles2021convergence,wang2020federated,wang2020tackling,wang2021field, acar2021federated}, and the impact of heterogeneity on FL has been explored more generally in connection with issues of fairness, robustness, and privacy in federated networks~\cite[e.g.,][]{mohri2019agnostic,li2020fair,li2021ditto,yu2020salvaging}.

The presumed presence of heterogeneity has also directly driven a large body of work in \textit{personalized federated learning}. In personalized FL, the goal is to learn or adapt models to more closely reflect the distinct data distribution of each federated client. For example, personalized FL approaches may consider fine-tuning a global or meta-trained model to adapt to local client data~\cite{zhao2018federated,jiang2019improving,chen2018federated,khodak2019adaptive,fallah2020personalized,singhal2021federated}, learning a clustering structure amongst the clients~\cite{ghosh2020efficient,sattler2020clustered,muhammad2020fedfast}, or using multi-task learning techniques to model possibly more complex relationships~\cite{smith2017federated,hanzely2020federated,hanzely2020lower,dinh2020personalized,li2021ditto}.

Despite a significant amount of work in personalized FL, there remains a lack of consensus about which methods perform best in various federated settings, let alone the degree to which personalization itself is really necessary in practice. For example, many works target a specific form of personalization (e.g., clustering) in their methodology, and then artifically create or modify data to match this assumption (e.g.,  manually clustering the data) before testing the efficacy of their approach~\cite{wang2021field}. While this is a reasonable sanity check, it fails to validate how impactful such approaches are for real-world FL applications. In particular, a major issue is that while `heterogeneity' is believed to be a natural occurrence in federated networks, there is a lack of benchmarks that reflect real applications of FL, making it hard to understand the prevalence and magnitude of heterogeneity in practice.

To address these concerns, we propose \name, a comprehensive benchmark for personalized and heterogeneity-aware FL. \name is designed with a focus on ease-of-use and reproducibility\footnote{\label{code}Code for the benchmark is open-source and available at: \codeurl.}. The benchmark includes a suite of four cross-device and three cross-silo datasets, as well as baseline personalization methods evaluated on this data. The datasets themselves are drawn from real-world applications mirroring federated settings in an effort to better reflect naturally occurring forms of heterogeneity. In developing baselines for the benchmark, we pay careful attention to the evaluation of personalization and heterogeneity---making concrete suggestions for future work in evaluating the impact of heterogeneity/personalization. Finally, although we focus specifically on the application of FL, we note that the benchmark is also a useful tool for the areas of transfer learning, meta learning, and multi-task learning more generally, as techniques from these areas are commonly used for personalized FL. The remainder of the paper is organized as follows:

\begin{itemize}[leftmargin=*]
    \item In Section~\ref{sec:background}, we provide background on federated learning and personalization. We make a key distinction between applications of cross-device vs. cross-silo FL, and highlight the ramifications of these differences in terms of methods that are most suitable for personalization (Table~\ref{table:cross-device-vs-cross-silo}).
    \item In Section~\ref{sec:benchmark}, we describe the components of the benchmark \name: (1) a diverse set of seven federated datasets and tasks chosen from public datasets covering both cross-device and cross-silo settings (Table~\ref{table:benchmark-overview}); (2) five representative personalized FL algorithms implemented in a modular fashion; and (3) baseline results across various evaluation metrics (Table~\ref{table:cross-device},  Table~\ref{table:cross-silo}). 
    \item Motivated by the benchmark, in Sections~\ref{sec:cross-device-exp} and~\ref{sec:cross-silo-exp} we provide suggestions for evaluating personalized FL and discuss practical concerns around using personalization in  real-world federated settings that are often overlooked in  existing literature. Our results highlight key insights about existing work such as the fundamental challenge of tuning hyperparmeters for personalization (Figure~\ref{fig:clients-hurt},  Figure~\ref{fig:tune_ft_hparams}), and point to several critical open directions of work in personalized FL.
    \item Finally, in Section~\ref{sec:conclusion}, we summarize our major take-aways from the benchmark, and point out limitations and future directions to expand benchmarking in this area more generally.
\end{itemize}

\textbf{Existing FL benchmarks.} Concurrent works benchmarking personalized FL~\cite{chen2022pfl, matsuda2022empirical} do not take into account differences between cross-device and cross-silo federated settings  (Table~\ref{table:cross-device-vs-cross-silo}), which we find to have a significant impact on the baseline methods and resulting conclusions. Beyond these works, other prior FL benchmarks (see, e.g, \cite{caldas2018leaf, lai2021fedscale, he2020fedml, he2021fedgraphnn, lee2020federated, chai2020fedeval, hu2020oarf, bouraqqadipyfed, beutel2020flower}) do not consider personalization baselines and focus instead on the standard FedAvg~\cite{mcmahan2017communication} algorithm and variations~\cite{reddi2021adaptive}. 

\section{Background: Federated Learning and Personalization}
\label{sec:background}

Federated learning (FL) considers the problem of learning across a set of distributed clients (e.g., sensors, phones, organizations). Based on the size and characteristics of the network, there are two common FL settings: \textit{cross-device FL} and \textit{cross-silo FL}, and two types of algorithms: \textit{stateful} and \textit{stateless}. We provide a brief description below, and defer readers to~\cite{kairouz2021advances} for more detailed discussion.

\textbf{Cross-device vs. Cross-silo.} Despite many similarities, it is worth distinguishing cross-device FL from cross-silo FL when understanding heterogeneity and the role of personalization. Cross-device applications typically involve learning across a large number (e.g., hundreds of millions) of mobile or IoT devices. Given the network scale coupled with the unreliability of such devices\footnote{\label{train-constraint}Usually only mobile phones that are idle, charging and connected to WiFi can join training~\cite{bonawitz2019towards}.}, it is common for devices to only participate once or \textit{never} participate in the training period~\cite{kairouz2021advances, wang2021field, yuan2021we}. This characteristic motivates the development of methods for cross-device FL which are \textit{stateless}\footnote{\label{stateless}Stateful algorithms can perform poorly in the cross-device settings where the clients sampling rate is very low in each round. Because most clients are sampled at most once during the entire training process, their local states are either unused or very stale. See Section 5.1 of~\cite{reddi2021adaptive} for the empirical results of the stateful SCAFFOLD algorithm~\cite{karimireddy2020scaffold} in the cross-device setting.}, in that model or variable state is not maintained on each client from one round to another, and there exists no unique identifier for each client (see more detailed discussions in~\cite{kairouz2021advances,wang2021field}). 

In contrast, cross-silo FL applications often consider learning across a handful of organizations such as hospitals or schools, where clients are almost always available at each training round. These properties allow the silos to be more easily accessed and identified, permitting the use of possibly more complex \textit{stateful} approaches. As we will see, these differences may change both what heterogeneity looks like in cross-device vs. cross-silo FL as well as the methods that are most appropriate to capture this heterogeneity~\cite{kairouz2021advances,wang2021field}. We therefore include examples of both cross-device and cross-silo FL datasets in the benchmark and consider methods and evaluation schemes unique to each setting. Table~\ref{table:cross-device-vs-cross-silo} summarizes the three key differences in the experimental setups between the cross-device and cross-silo settings.

\begin{table}[ht]
\caption{Differences between the cross-device and cross-silo experimental setups.}
    \centering
    \begin{tabular}{p{5.2cm}p{4.9cm}p{5.3cm}}
    \toprule
    \textbf{Experimental setup} & \textbf{Cross-device} & \textbf{Cross-silo}\\
    \midrule
    Client sampling rate per training round (Section~\ref{sec:benchmark-data}) & Typically on the scale of 0.1\%-1\% (see Table~\ref{table:cross-device-sampling} for the exact value) & 100\% (all silos participate in every round) \\
    \midrule
    Train/valid/test split (Section~\ref{sec:benchmark-data}) & Split the clients (see Fig.~\ref{fig:data-hetero}) & Split each client's data (see Fig.~\ref{fig:data-hetero}) \\
    \midrule
    Stateful or stateless algorithm & Stateless\footnoteref{stateless}. & Both work. \\
    \bottomrule
    \end{tabular}
    \label{table:cross-device-vs-cross-silo}
\end{table}

\textbf{Heterogeneity\footnote{\label{hetero}Note that we focus on \textit{data heterogeneity} here and do not consider other types of heterogeneity in our experiments, e.g., mobile devices may have different processing speeds and memory constraints, which is worth exploring in the future.}.} Motivated by privacy and efficiency concerns, in both cross-device and cross-silo FL it is common to keep the raw data generated on each distributed client local to each client, and instead aggregate focused updates to perform collaborative learning~\cite{mcmahan2017communication}. While this has many practical benefits, it can also introduce or exacerbate issues of heterogeneity, as each client may generate data according to a unique distribution. To better account for this heterogeneity, it is therefore increasingly common to consider techniques (described below) that learn personalized, client-specific models.

\textbf{Personalization.} We defer readers to Appendix~\ref{app:background} and the recent surveys~\cite{tan2022towards}, \cite[\S7.5]{wang2021field} for more detailed discussion on existing literature in personalized FL. In general, existing personalized FL approaches can be categorized in three different ways:
\begin{itemize}[leftmargin=*]
    \item Stateful vs stateless: a statuful algorithm~\cite[e.g.,][]{karimireddy2020scaffold, smith2017federated,huang2021personalized,li2021ditto,sattler2020clustered, liang2020think,arivazhagan2019federated, hanzely2021personalized, marfoq2021federated} requires certain states to be maintained on participating clients from one round to another (e.g., SCAFFOLD algorithm~\cite{karimireddy2020scaffold} requires each client maintain a control variate locally\footnoteref{stateless}). As pointed out in Table~\ref{table:cross-device-vs-cross-silo}, while stateless approaches are applicable to both settings, stateful approaches are more appropriate for cross-silo FL (see also~\cite{kairouz2021advances,wang2021field}). 
    \item Model-agnostic vs model-specific: whether the method targets a specific model architecture or requires domain-specific information~\cite[e.g.,][]{singhal2021federated, zhu2021diurnal, jain2021differentially, li2021fedbn}).
    \item In terms of the methodologies: e.g., meta-learning~\cite[e.g.,][]{jiang2019improving, wang2019federated, fallah2020personalized,khodak2019adaptive,charles2021convergence, acar2021debiasing}, clustering~\cite[e.g.,][]{sattler2020clustered, mansour2020three, ghosh2020efficient, xie2020multi}, local memory~\cite{marfoq2022personalized}, and multi-task learning (MTL)~\cite[e.g.,][]{hanzely2020federated,hanzely2020lower,dinh2020personalized,li2021ditto,smith2017federated}.
\end{itemize}
In benchmarking \name, we consider five model-agnostic personalized FL methods (four stateless and one stateful) from the major methodology categories (meta-learning, clustering, local memory, and MTL).

\section{\name: A Benchmark for Personalized Federated Learning}\label{sec:benchmark}

Table~\ref{table:benchmark-overview} gives an overview of the three components of \name (methods, datasets, baselines). We describe the methods and datasets in this section. In Sections~\ref{sec:cross-device-exp} and~\ref{sec:cross-silo-exp}, we provide baseline results and discuss the practical concerns of performing personalization in real-world FL applications.

\subsection{Personalization Methods}\label{sec:benchmark-algo}

\name contains five simple and model-agnostic algorithms for learning personalized models. The first four algorithms are \emph{stateless}, and hence, are appropriate in the cross-device FL setting (see Table~\ref{table:cross-device-vs-cross-silo}). For all the experiments, we use the generalized version of the FedAvg algorithm~\cite{reddi2021adaptive}, in which the server optimizer can use adaptive optimizers. This generalizes the original FedAvg algorithm~\cite{mcmahan2017communication} where the server only averages the client updates.

\begin{itemize}[leftmargin=*]
\item \textbf{Local training} refers to every client training a local model using their own data, without collaborating with others. While it is not an FL algorithm, we include it in \name as it provides a good baseline for FL algorithms, and it may be competitive when clients have sufficient local data~\cite{yu2020salvaging}.   
\item \textbf{FedAvg+Fine-tuning}~\cite{jiang2019improving} is a simple method for stateless, model-agnostic personalized FL, operating as follows: First train a global model via FedAvg~\cite{reddi2021adaptive}; then, each client fine-tunes the global model on their local data and uses the fine-tuned model for inference. It has a natural connection to meta-learning~\cite{jiang2019improving,khodak2019adaptive}\footnote{Specifically, FedAvg can be viewed as performing the Reptile algorithm~\cite{nichol2018first} to learn an effective starting model, such that after fine-tuning, the model can quickly adapt to a client's local data.}, and has been reported to work well in real-world on-device applications~\cite{wang2019federated,sim2021robust}. In \name, we explore two variations: fine-tuning the entire model or fine-tuning only the last layer (the latter can be viewed as a form of hard parameter sharing MTL). 
\item \textbf{HypCluster}~\cite{mansour2020three} (also \textbf{IFCA}~\cite{ghosh2020efficient}) is a stateless method that jointly clusters clients and learns a model for each cluster. Both HypCluster and IFCA work as follows: in each training round, the server sends all models (one per cluster) to the participating clients; each client receives the models and picks the cluster associated to the model with the lowest loss on its local data. It then computes an update for the selected model and sends the update and cluster identity to the server. The server aggregates the model updates for each cluster in the same way as in FedAvg~\cite{reddi2021adaptive}. In \name, we explore two initialization strategies: random and warm start with FedAvg~\cite{reddi2021adaptive}.
\item \textbf{FedAvg+kNN-Per}~\cite{marfoq2022personalized} is a recent personalization algorithm. The idea is to interpolate/ensemble the output of two models: a globally-trained FedAvg model and a local k-Nearest Neighbors (kNN) model. For each client's local example, a representation vector is obtained from the FedAvg model (e.g., the input to the last softmax layer, or the states of an LSTM model). The Euclidean distance between the representations are used to learn a kNN model.
\item \textbf{Multi-Task Learning} (MTL) is a class of methods used to deliver personalized models for a set of tasks by learning the task relations (either explicitly or implicitly). Each `task' corresponds to a client in the FL setting. MTL approaches usually require the clients to be stateful~\citep[e.g.,][]{hanzely2020federated,smith2017federated}, and hence, are more appropriate for cross-silo settings (see Section~\ref{sec:background} on stateless vs stateful algorithms). Many existing personalized FL methods can be viewed as a form of MTL~\citep[e.g.,][]{hanzely2020federated,hanzely2020lower,dinh2020personalized,li2021ditto,smith2017federated}. In \name, we consider two MTL algorithms: 1) Mocha~\cite{smith2017federated}, the first work that proposes to personalize federated models in convex settings, and 2) Ditto~\cite{li2021ditto}, a recent work that regularizes personalized models towards optimal global models for both convex and non-convex problems.
\end{itemize}

\begin{table}[t]
  \caption{\name has three components: (1) modular implementations of five representative personalized FL algorithms, (2) a diverse range of tasks (C: classification; R: regression; NWP: next word prediction) and datasets (see Appendix~\ref{app:data}) chosen to cover the cross-device and cross-silo FL settings, and (3) baseline results via extensive hyparameter tuning (see Appendix~\ref{app:hparam}) and the insights obtained from these results. The right-most column is the average number of examples per client $\pm$ standard deviation.}
  \label{table:benchmark-overview}
  \centering
  \begin{tabular}{@{}p{4.6cm}p{3cm} p{3.5cm} p{1.2cm} p{1.5cm}}
    \toprule
    \centering{\textbf{Methods}} & \multicolumn{4}{c}{\textbf{Dataset Details}}\\
    & \textit{Dataset} & \textit{Task and Model} & \textit{Clients} & \textit{Pts/Client}\\
    \midrule
    \textit{Cross-Device FL} \\
  \multirow{5}{4.6cm}{\hspace{1em}Local training\\\hspace{1em}FedAvg+Fine-tuning~\cite{jiang2019improving}\\\hspace{1em}HypCluster~\cite{mansour2020three}/IFCA~\cite{ghosh2020efficient}\\\hspace{1em}FedAvg+kNN-Per~\cite{marfoq2022personalized}}\\ & EMNIST~\cite{emnist} & Image C; CNN & 3400 & 198$\pm$89 \\
     & StackOverflow~\cite{stackoverflow} & NWP; LSTM & 380k & 397$\pm$1279   \\ 
    & Landmarks~\cite{hsu2020federated} & Image C; MobileNetV2 & 1262 & 130$\pm$199 \\ 
    & TedMulti-EnEs~\cite{qi2018and} & NWP; Transformer &4184&113$\pm$56\\
    \midrule
    \textit{Cross-Silo FL} \\
    \hspace{1em}Local training & ADNI\footnoteref{adni} & Image R; CNN & 9 & 5405$\pm$4822 \\ 
    \hspace{1em}FedAvg+Fine-tuning~\cite{jiang2019improving} & Vehicle~\cite{duarte2004vehicle} & Binary C; SVM & 23 & 1900$\pm$349 \\ 
    \hspace{1em}HypCluster~\cite{mansour2020three}/IFCA~\cite{ghosh2020efficient} & School\footnoteref{school} & R; Linear Regression & 139 & 111$\pm$56 \\ 
    \hspace{1em}FedAvg+kNN-Per~\cite{marfoq2022personalized} & &&&\\
    \hspace{1em}Multi-task learning~\cite{smith2017federated, li2021ditto} & &&&\\
    \bottomrule
  \end{tabular}
\end{table}

\begin{figure}[ht]
    \centering
    \begin{subfigure}[b]{0.48\textwidth}
    \includegraphics[width=\textwidth]{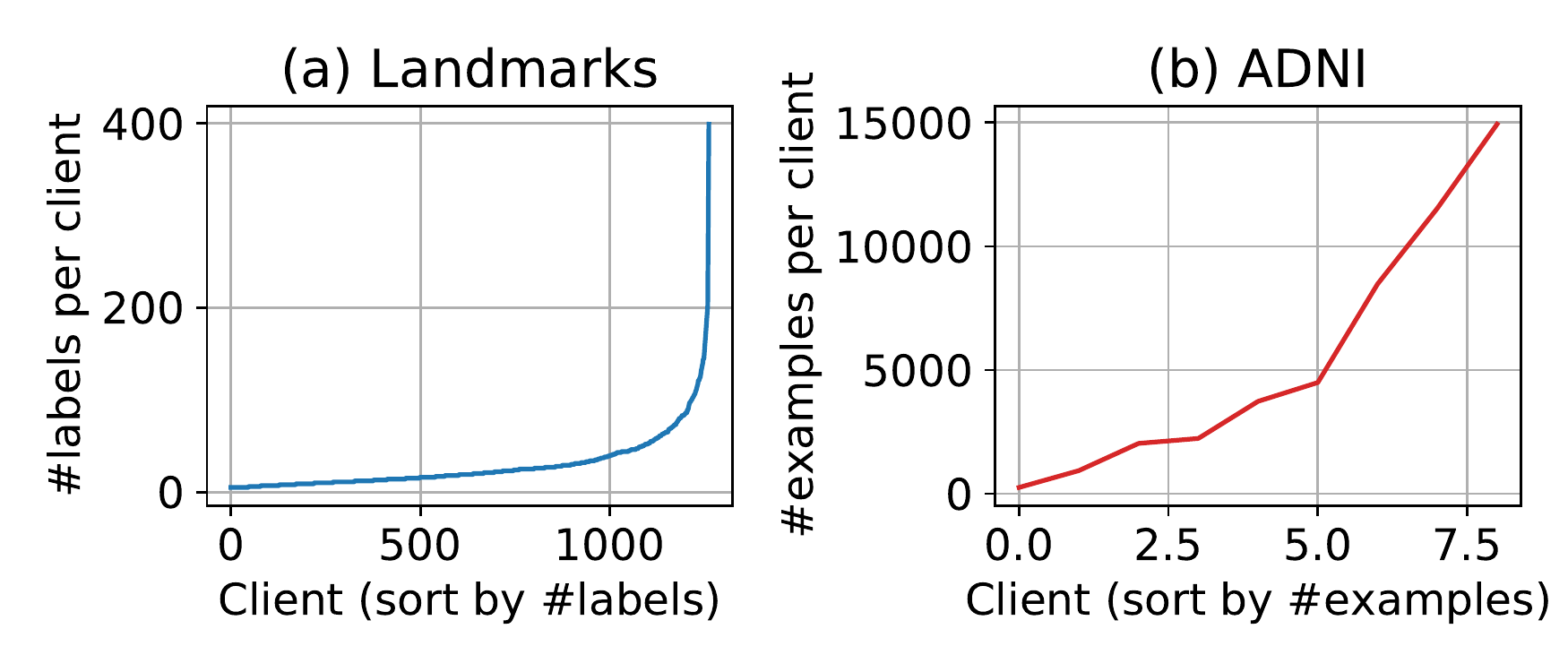}
    \end{subfigure}
    \hfill
    \begin{subfigure}[b]{0.5\textwidth}
    \includegraphics[width=\textwidth]{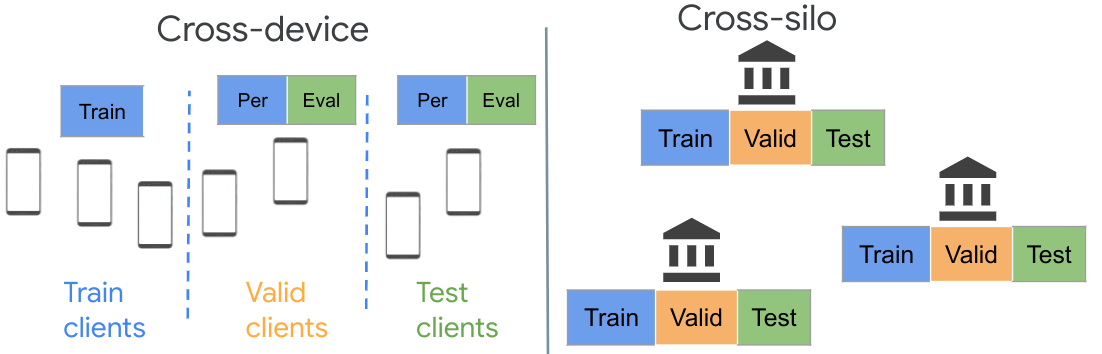}
    \end{subfigure}
    \caption{\textbf{Left}: The chosen federated datasets (Table~\ref{table:benchmark-overview}) have heterogeneous local distributions (see Appendix~\ref{app:data} for other datasets). \textbf{Right}: To best reflect real-world FL applications, we naturally pre-process cross-device and cross-silo datasets differently: in cross-device, we split the clients into train/validation/test, and each validation/test client's local examples are split into two local sets: a personalization set and an evaluation set; in cross-silo, we split each client's local dataset into train/validation/test sets (see Section~\ref{sec:benchmark-data}).}
    \label{fig:data-hetero}
\end{figure}

\subsection{Datasets and Pre-processing}\label{sec:benchmark-data}

Table~\ref{table:benchmark-overview} lists the datasets carefully chosen to reflect real-world FL applications, including a health data\footnote{\label{adni}Health data obtained from Alzheimer’s Disease Neuroimaging Initiative: \url{https://adni.loni.usc.edu/}.} and a school data\footnote{\label{school}Data collected by \url{https://en.wikipedia.org/wiki/Inner_London_Education_Authority}.} in the cross-silo setting (see Appendix~\ref{app:data} for detailed descriptions). All datasets have natural per-user partitions and distinct local statistics (Figure~\ref{fig:data-hetero}). \name provides data pre-processsing pipelines for all data\footnoteref{code}, and includes a critical and often overlooked distinction between pre-processing cross-device vs. cross-silo data (described below).

\begin{itemize}[leftmargin=*]
    \item \textbf{Cross-device}. As shown in Figure~\ref{fig:data-hetero}, we first randomly split the \emph{clients} into three disjoint sets: train, validation (for hyperparameter tuning), and test (for final evaluation). This split reflects practical cross-device FL settings: given the population scale (e.g., millions of mobile devices~\cite{hard2018federated}) and that clients usually need to satisfy certain conditions such as idle, WiFi, and charging in order to be able to join training~\cite{bonawitz2019towards}, devices participating in inference may never participate in training. It is therefore important to ensure that the trained models can generalize to clients that are unseen during the training time~\cite{yuan2021we}. We then split each validation and test client's local examples into two equal-sized sets: a personalization set (for learning a personalized model, e.g., via fine-tuning, building a kNN model, or choosing best model among multiple models) and an evaluation set (for evaluating the personalized model)\footnote{On StackOverflow, we \emph{sort} the local examples by time before splitting so that the model is fine-tuned on the old examples and evaluated on the new ones, reflecting practical FL settings. We perform random split on other datasets because they do not save time information. This means that, except StackOverflow, our experiments do not capture the potential distribution shift between the old and new examples. See Section~\ref{sec:cross-device-exp-FT} for why this can potentially benefit FedAvg+Fine-tuning.}. See Section~\ref{sec:cross-device-exp} for how these datasets are used in each algorithm.  
    \item \textbf{Cross-silo.} Unlike cross-device FL, in the cross-silo FL setting the total number of silos is small (e.g., tens of hospitals), and the same silos usually participate in both training \emph{and} inference. To evaluate this setting, we split each silo's local examples into three sets: a train, validation, and test set (as shown in Figure~\ref{fig:data-hetero}). See Section~\ref{sec:cross-silo-exp} for details on how they are used in training and evaluation.
\end{itemize}

\textbf{Clients sampling.} Besides the distinction in the pre-processing steps, cross-device and cross-silo settings have very different clients sampling rates in training. In the real-world cross-device applications, the clients sampling rate is typically very low~\cite{kairouz2021advances, wang2021field}, e.g., the total population is hundreds of millions devices but only a few thousands participate at every training round. Table~\ref{table:cross-device-sampling} lists the number of clients sampled per round used in our cross-device experiments. In the cross-silo experiments, we assume that all clients are always available at each training round as shown in Table~\ref{table:cross-device-vs-cross-silo}.

\begin{table}[ht]
\caption{Number of clients sampled per training round in our cross-device experiments (see Appendix~\ref{app:hparam} for all the hyperparameters). For EMNIST, StackOverflow, and Landmarks, we use the same number of sampled clients per round as in~\cite{charles2021large, hsu2020federated}. The real-world cross-devices settings usually have a very low sampling rate, e.g., hundreds of millions devices in total but only a few thousands participate each round~\cite{kairouz2021advances}.}
    \centering
    \begin{tabular}{p{2.6cm}p{3.8cm}p{4.2cm}p{2.6cm}} \toprule
    \textbf{Datasets} & \textbf{Total train clients} & \textbf{Sampled clients/round} & \textbf{Sampling rate}\\
    \midrule
    EMNIST & 2500 & 50 & 2\% \\
    StackOverflow & 342,477 & 200 & 0.05\% \\
    Landmarks & 1112 & 64 & 6\% \\
    TedMulti-EnEs & 3969 & 32 & 0.8\% \\
    \bottomrule
    \end{tabular}
    \label{table:cross-device-sampling}
\end{table}

\section{Cross-Device Experiments}\label{sec:cross-device-exp}

We now provide the cross-device personalization baselines by evaluating the four stateless methods described in Section~\ref{sec:benchmark} and discuss our findings. The results are summarized in Table~\ref{table:cross-device}.

\begin{table}[ht]
  \renewcommand{\arraystretch}{1.1}
  \caption{Summary of experimental results on the cross-device FL datasets as well as possible concerns around using each method in  practice. We report the per-client accuracy (mean $\pm$ standard deviation) on the test clients after training 1500 rounds (following \cite{charles2021large}) on EMNIST, StackOverflow and TenMulti, and 30k rounds (following \cite{hsu2020federated}) on Landmarks. Note that the standard deviation is across all the clients' local accuracies, which is a fairness metric considered in~\cite{mohri2019agnostic, li2021ditto}. Each value is further averaged over 5 different runs (see Table~\ref{apptable:cross-device} for the standard deviation over the 5 runs). Appendix~\ref{app:hparam} provides the tuned hyperparameters.}
  \label{table:cross-device}
  \centering
  \begin{tabular}{p{2.2cm}|p{3.8cm}p{1.9cm}p{2.3cm}p{1.9cm}p{1.9cm}} \toprule
    \textbf{Algorithm} & \textbf{Metrics} & \textbf{EMNIST} & \textbf{StackOverflow} & \textbf{Landmarks} & \textbf{TedMulti}\\
    \midrule
    Local training & Per-client acc & 0.594$\pm$.17 &0.062$\pm$.03&0.173$\pm$.16&0.056$\pm$.02 \\
    \midrule
    \multirow{6}{2.2cm}{FedAvg + Fine-tuning (FT)}&Per-client acc before FT & 0.844$\pm$.10 & 0.269$\pm$.03 & 0.564$\pm$.16 &0.160$\pm$.04\\
    &Per-client acc after FT & 0.903$\pm$.06 & 0.282$\pm$.03& 0.773$\pm$.11 &0.162$\pm$.04 \\
    &\% clients "hurt" after FT &5.2\%&14\%&5.6\%&40\%\\
    &FT all layers vs last layer & All layers & All layers & All layers &All layers\\ 
    \cmidrule{2-6}
    &\multicolumn{5}{c}{\textit{\textbf{Practical concerns}: difficult to tune hyperparameters; may hurt clients; sensitive}}\\
    &\multicolumn{5}{c}{\textit{to distribution shift; performance drops with fewer local examples (see Section~\ref{sec:cross-device-exp-FT})}}\\
    \midrule
    \multirow{8}{2.2cm}{HypCluster / IFCA}&Per-client acc & 0.897$\pm$.08 &0.273$\pm$.03& 0.573$\pm$.16 &0.163$\pm$.04\\
    & No. tuned clusters ($k$)& 2 & 2 & 2 & 2\\
    & \% clients largest cluster & 52.6\% &85.1\%&92.1\%&54.7\%\\
    & Warmstart from FedAvg & Yes & Yes & Yes&Yes\\
    & Per-client acc by ensembling $k$ FedAvg models & 0.860$\pm$.08&0.271$\pm$.03&0.564$\pm$.16&0.163$\pm$.04\\ \cmidrule{2-6}
    &\multicolumn{5}{c}{\textit{\textbf{Practical concerns}: difficult to train due to mode collapse; high communication cost;}}\\
    &\multicolumn{5}{c}{\textit{difficult to interpret the learned clusters; sensitive to distribution shift (see Section~\ref{sec:cross-device-exp-cluster})}}\\
    \midrule
    \multirow{4}{2.2cm}{FedAvg + kNN-Per}&Per-client acc & 0.876$\pm$.06 & 0.275$\pm$.03 & 0.735$\pm$.13 & 0.162$\pm$.05\\
    &\% clients "hurt"
    &19.8\%& 23.6\%&5.6\%&34.4\%\\
    \cmidrule{2-6}
    &\multicolumn{5}{c}{\textit{\textbf{Practical concerns}: difficult to tune hyperparameters; may hurt clients; sensitive}}\\
    &\multicolumn{5}{c}{\textit{to distribution shift; performance drops with fewer local examples (see Section~\ref{sec:cross-device-exp-knn})}}\\
    \bottomrule
    \end{tabular}
\end{table}

\begin{figure}[ht]
    \centering
    \begin{subfigure}[b]{0.24\textwidth}
    \includegraphics[width=\textwidth]{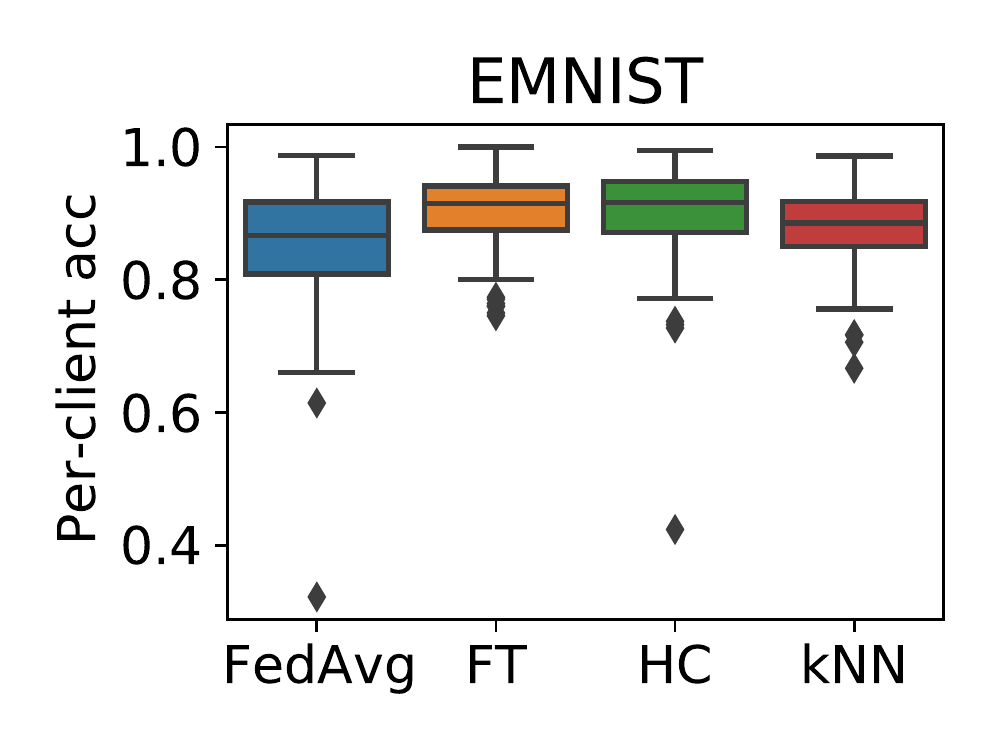}
    \end{subfigure}
    \hfill
    \begin{subfigure}[b]{0.24\textwidth}
    \includegraphics[width=\textwidth]{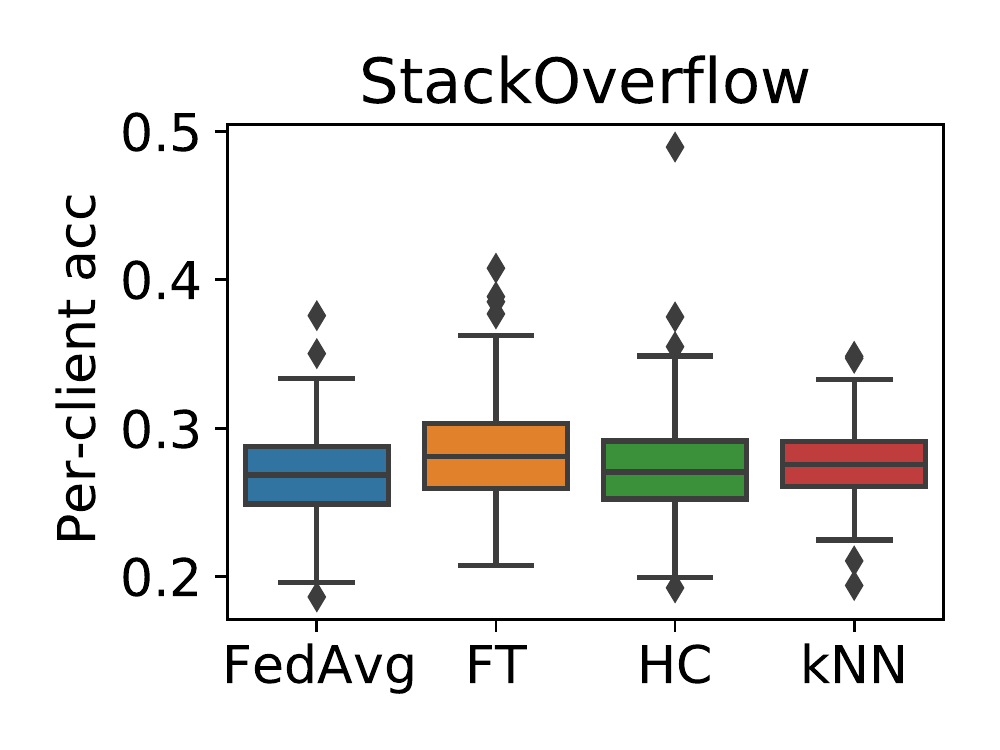}
    \end{subfigure}
    \hfill
    \begin{subfigure}[b]{0.24\textwidth}
    \includegraphics[width=\textwidth]{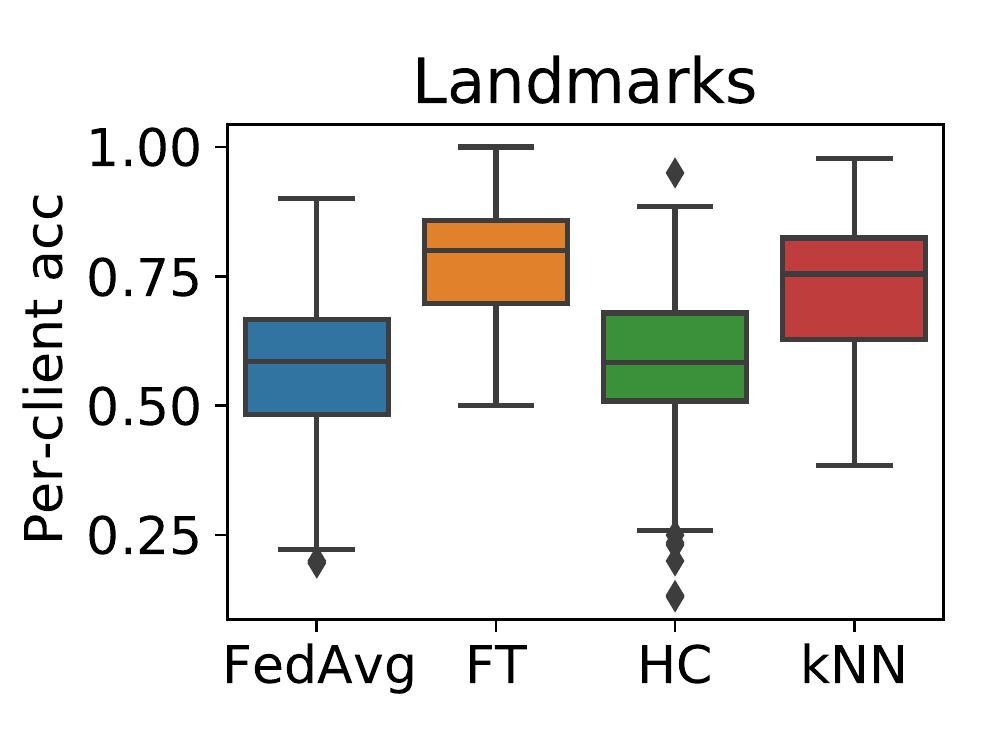}
    \end{subfigure}
    \hfill
    \begin{subfigure}[b]{0.24\textwidth}
    \includegraphics[width=\textwidth]{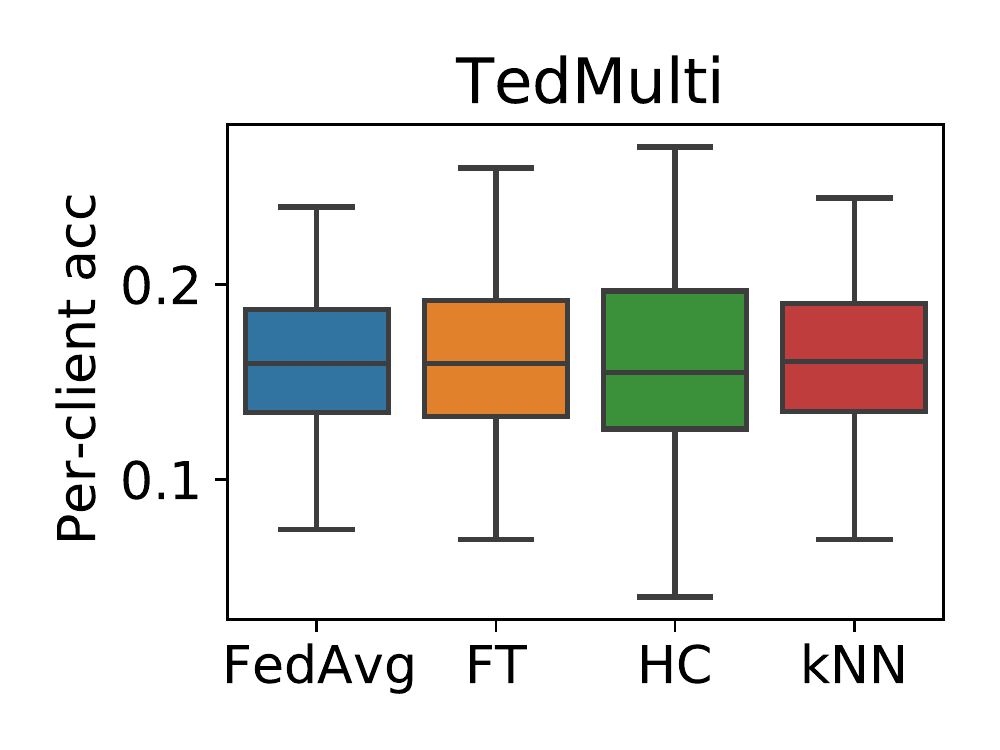}
    \end{subfigure}
    \caption{Test clients' per-client accuracy shown in box plots. Fine-tuning gives the best average per-client accuracy on EMNIST, StackOverflow, and Landmarks (explicit values are in Table~\ref{table:cross-device}).}
    \label{fig:box-plot}
\end{figure}

%

\textbf{Training and evaluation process.} As mentioned in Section~\ref{sec:benchmark-data}, for each cross-device dataset, we split the total clients into the train, validation, and test clients. Besides, the validation clients and test clients' local data is split into two sets: a personalization set and an evaluation set. The train clients are used to train a single model via FedAvg or multiple models via HypCluster. The trained model(s) are then evaluated on the validation clients \emph{and} the test clients. 
\begin{itemize}[leftmargin=*]
    \item For FedAvg + Fine-tuning, this evaluation is done as follows: each (validation and test) client first fine-tunes the FedAvg-trained model on the local personalization set, then evaluates the fine-tuned model on the local evaluation set. 
    \item For HypCluster, this evaluation is performed as follows: each client uses the local personalization set to find the model (among the multiple models learned by HypCluster) with the lowest loss, and then evaluates the selected model on the evaluation set. 
    \item For FedAvg+kNN-Per, each client first uses the FedAvg-trained model to extract the representations on the local personalization set, and use them to train a kNN model; after that, evaluates the personalized model (interpolated between the kNN model and FedAvg-trained model) on the evaluation set.
    \item For local training, the evaluation process is same as FedAvg+Fine-tuning, except that each client fine-tunes a model from scratch instead of from a FedAvg-trained model.
\end{itemize}
We use the validation metrics (i.e., the metrics evaluated over the validation clients) to select the best hyperparameters (see Appendix~\ref{app:hparam}) and report the test metrics (i.e., metrics evaluated on the test clients) in Table~\ref{table:cross-device}. We discuss additional details and results for each stateless FL method below. 

\subsection{FedAvg + Fine-tuning (FT)}\label{sec:cross-device-exp-FT}
As we focus on personalization, it is crucial to see how the per-client accuracy changes with and without personalization. This is shown by the "Per-client acc before/after FT" metrics in Table~\ref{table:cross-device}, where we list the mean $\pm$ standard deviation of the test clients' per-client accuracy before and after fine-tuning. Note that the standard deviation is across the clients' local per-client accuracies, and is hence, an fairness metric~\footnoteref{fairness}. The mean and standard deviation metrics are further averaged over 5 runs\footnote{\label{std-5runs}See Appendix~\ref{app:more-exp} for the standard deviations across the 5 runs.}. 
As shown in Table~\ref{table:cross-device}, fine-tuning improves the average per-client accuracy on EMNIST, StackOverflow, and Landmarks (see also Figure~\ref{fig:box-plot}), and reduces the standard deviation across the clients' local model accuracies on EMNIST and Landmarks (i.e., improves \emph{fairness}\footnote{\label{fairness}There are many notions of fairness, e.g., group fairness \cite{hardt2016equality}. In this paper we use a notion specific to federated learning (see, e.g., \cite{li2021ditto, mohri2019agnostic}): all clients should have similar local model accuracies.}). We explored fine-tuning the entire model or only last layer and found that fine-tuning all layers perform better, as shown by the metric "FT all layers vs last layer" in Table~\ref{table:cross-device}.

\begin{figure}[ht]
    \centering
    \includegraphics[width=\linewidth]{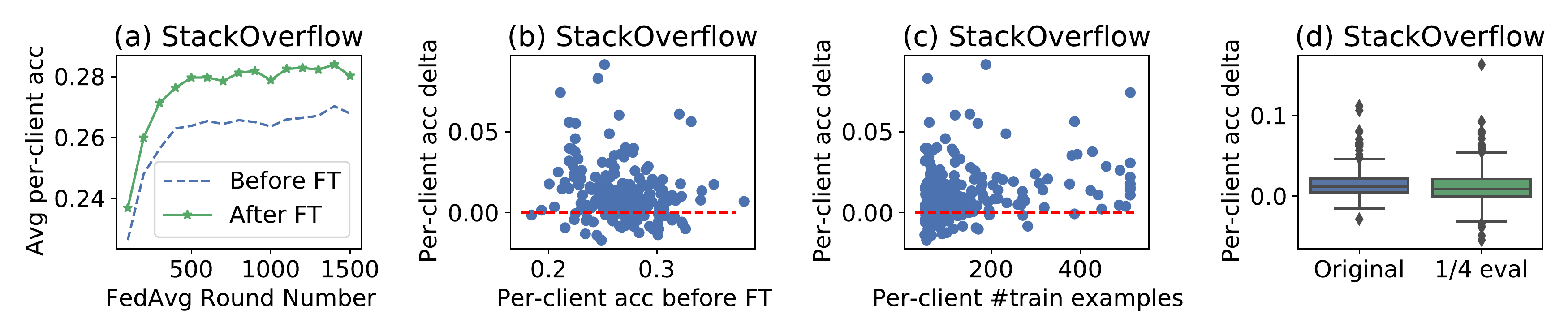}
    \caption{Fine-tuning may hurt some clients. (a) Fine-tuning improves the average per-client accuracy. (b) Scatter plot of per-client accuracy delta (i.e., accuracy after fine-tuning -- before fine-tuning). Each dot is a client. (c) Same scatter plot as (b) with the x-axis being the number of local examples used to fine-tune the model. (d) Fine-tune the same model on the same set of clients using smaller evaluation set.}
    \label{fig:clients-hurt}
\end{figure}

\textbf{Observation 1: Fine-tuning can hurt clients.} The metric "\% clients hurt after FT" in Table~\ref{table:cross-device} is the percentage of test clients whose local accuracy drops after fine-tuning. This phenomenon can be better visualized in Figure~\ref{fig:clients-hurt}: while Figure~\ref{fig:clients-hurt}(a) shows that fine-tuning increases the average per-client accuracy on the StackOverflow dataset, the scatter plot (each dot is a client) of the per-client accuracy deltas (defined as accuracy after fine-tuning minus that before fine-tuning) in Figure~\ref{fig:clients-hurt}(b) shows that some clients drop accuracy after fine-tuning (i.e., fall below the red dashed line at zero). We identify two reasons:
\begin{enumerate}[leftmargin={0.17in},label=\alph*.]
    \item The per-client accuracy is noisy for clients with small local datasets. This effect occurs in two distinct ways: 1) Clients use a small number of examples to fine-tune the model\footnote{\label{ft-thres}In practice, it may be good to only apply fine-tuning to clients when they have a large amount of local data.}. In Figure~\ref{fig:clients-hurt}(c), we show the same scatter plot (each dot is a client) as in Figure~\ref{fig:clients-hurt}(b) with the x-axis being the number of examples used to fine-tune the model. The clients hurt by fine-tuning tend to have a small local personalization set\footnote{\label{split}See Section~\ref{sec:benchmark-data}, each validation and test client has two sets: a personalization set and an evaluation set.}. 2) Clients use a small number of examples to evaluate the fine-tuned model. In Figure~\ref{fig:clients-hurt}(d), we fine-tune the same model on the same set of clients. The only difference is the size of the evaluation set\footnoteref{split}. The evaluation metrics are noisier (indicated by a larger range in Figure~\ref{fig:clients-hurt}(d)) for smaller evaluation sets.
    \item Heterogeneity among clients and the fact that the fine-tuning hyperparameters are chosen globally. Since each client's local dataset is typically small, instead of choosing the fine-tuning hyperparameters (i.e., the fine-tuning learning rate, the number of fine-tuning epochs, and which layer to fine-tune) in a per-client manner\footnote{\label{per-client-tune}We actually tried this approach (i.e., each client chooses the fine-tuning hyperparameters based on its local data) on the StackOverflow dataset, but found that it performed worse than tuning the hyperparameters globally.}, we tune them globally and apply the same hyperparameters to all test clients. The global fine-tuning hyperparameters work for most clients, but can adversely affect others. This also illustrates why hyperparameter tuning is difficult (see below).
\end{enumerate}

\begin{figure}
    \centering
    \includegraphics[width=\linewidth]{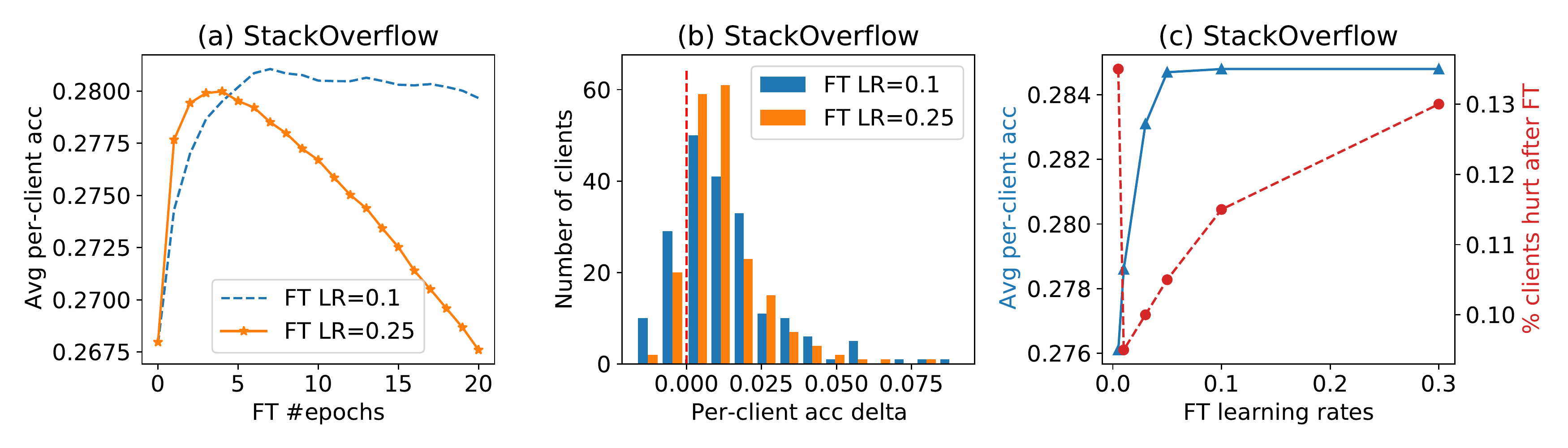}
    \caption{Choosing a good set of fine-tuning hyperparameters is difficult, as different metrics may be improved by different hyperparameters. (a) FT learning rate 0.1 gives higher average per-client accuracies; (b) FT learning rate 0.25 hurts fewer clients; (c) The "avg per-client acc" metric may prefer a moderately large FT learning rate, while the "clients hurt" metric prefers a small FT learning rate.}
    \label{fig:tune_ft_hparams}
\end{figure}

\textbf{Observation 2: Hyperparameter tuning can be difficult.} At least three hyperparameters are specific to fine-tuning: which layers to fine-tune, the fine-tuning learning rate\footnote{\label{ft-sgd}We focus on SGD when fine-tuning the model on each client, as adaptive optimizers seem to perform similarly as SGD.}, and the number of epochs to fine-tune the model. As pointed out in Observation 1(b), due the small local dataset on each client, instead of choosing the best fine-tuning hyperparameters in a per-client manner, we typically tune them globally and apply the same hyperparameters to all the test clients\footnoteref{per-client-tune}. Choosing a good set of fine-tuning hyperparameters can be particularly difficult due to the following  reasons: 
\begin{enumerate}[leftmargin={0.17in},label=\alph*.]
    \item Different metrics may favor different hyperparameters\footnote{The results shown in Table~\ref{table:cross-device} is based on tuning hyperparameters by the average per-client accuracy.}. For example, if we look at Figure~\ref{fig:tune_ft_hparams}(a), we may want to choose 0.1 as the fine-tuning learning rate because it gives higher average per-client accuracy; however, if we look at Figure~\ref{fig:tune_ft_hparams}(b), we may want to choose 0.25 because fewer clients drop their accuracies after fine-tuning (i.e., per-client acc delta < 0). In Figure~\ref{fig:tune_ft_hparams}(c), we plot the two metrics across different fine-tuning learning rates: the "average per-client accuracy"metric prefers a moderately large learning rate while the "clients hurt" metric prefers a small learning rate.
    \item Since the per-client accuracies are noisy, we need to be careful when determining whether the difference between two sets of hyperparameters is statistically significant.
    \item Extra hyperparameters add another layer of complexity. For example, it may be helpful to tune FedAvg together with fine-tuning so that the model produced by FedAvg is a good initial model for fine-tuning~\cite{jiang2019improving, khodak2021federated}. 
    Another example of extra hyperparameters is that one may want to learn a threshold\footnoteref{ft-thres} such that only clients whose local dataset size is larger than this threshold can perform fine-tuning (see also the remark on "acceptance criteria and robust personalization" after Observation 4).  
\end{enumerate}

\begin{figure}
    \centering
    \includegraphics[width=\linewidth]{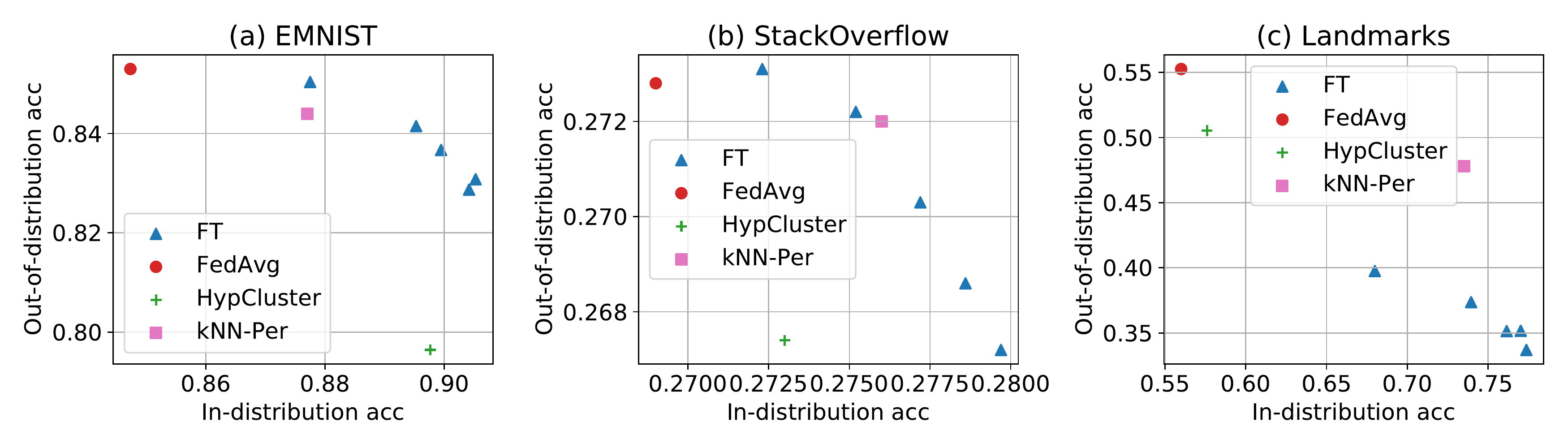}
    \caption{Trade-off between in-distribution (ID) and out-of-distribution (OOD) accuracy for different algorithms. We plot the results for fine-tuning epochs $1,3,5,10,15$. Increasing the fine-tuning epochs gives higher ID accuracy but lower OOD accuracy.}
    \label{fig:id_ood_tradeoff}
\end{figure}

\textbf{Observation 3: Sensitivity to distribution shift.} Fine-tuning is able to quickly adapt a model to a client's local data, but the fine-tuned model may perform poorly if a client's future data distribution differs from the existing data (a.k.a. catastrophic forgetting~\cite{mccloskey1989catastrophic}). To illustrate this, we evaluate each test client's personalized model over the in-distribution (ID) test set (i.e., the client's local evaluation set\footnoteref{split}) and an out-of-distribution (OOD) test set. This OOD test set is formed by randomly sampling local examples from the entire test clients. The averaged ID and OOD test accuracies are shown in Figure~\ref{fig:id_ood_tradeoff}. We plot the results for different fine-tuning epochs. Larger fine-tuning epoch gives higher ID accuracy but lower OOD accuracy. This trade-off indicates an extra maintenance cost of personalized algorithms, i.e., decide when and how to continuously personalize the model when clients have new data.

\textbf{Observation 4: Performance drops when clients have smaller personalization set\footnoteref{split}.} This is already indicated previously in Figure~\ref{fig:clients-hurt}(c), which shows that clients with smaller personalization set (i.e., fewer examples used to fine-tuned the model) tend to be hurt by fine-tuning. In Figure~\ref{fig:knn_per_hparams}(c), we reduce the number of examples in each test client's persoanlization set to 50\% and 25\% of its original size, and report the per-client accuracy averaged over the test clients. We see that the performance of fine-tuning and kNN-Per drops quickly when we decrease the number of local examples to personalize the model. On the other hand, HypCluster is quite robust to this change, which makes sense because HypCluster only needs the personalization set for model evaluation and use the evaluation metrics to choose the best model, while fine-tuning and kNN-Per needs the local examples to train a new model.  

\textbf{Remark: Acceptance criteria and robust personalization.} Our experiments assume that every client always performs fine-tuning and uses the fine-tuned model for inference. In practice, one can design appropriate criteria so that a client only accepts the fine-tuned model under certain conditions. For example, a client only performs fine-tuning when its local dataset size is large enough as discussed in Observation 4; or a client only accepts the fine-tuned model when its evaluation metric is better than that of the baseline model by some margin. See Section 2.2 of~\cite{sim2021robust} for an example of acceptance criteria developed for a real-world on-device personalization application. Designing a good set of acceptance criteria can mitigate the negative effects from Observation 1--4 above, and potentially prevent clients from accepting fine-tuned models that do not generalize well. Mitigating catastrophic forgetting~\cite{mccloskey1989catastrophic} during model adaptation is an active research area, see, e.g., some recent work~\cite{andreassen2021evolution, kumar2022fine, wortsman2022robust, wortsman2022model, ilharco2022patching}, which would be worth exploring in the personalized FL setting.

\subsection{HypCluster / IFCA} \label{sec:cross-device-exp-cluster}

We now discuss results of HypCluster/IFCA algorithm. As shown in Table~\ref{table:cross-device}, the average per-client accuracy of HypCluster is slightly better than that of FedAvg, but is much worse than that of FedAvg+Fine-tuning on the first three datasets. We tuned\footnote{Except for TedMulti-EnEs which we focus on 2 clusters since it contains two languages: English/Spanish.} the number of clusters $k=\{2,3,4\}$; however, due to the mode collapse issue (described below), it is difficult to train HypCluster with $k>2$. A natural baseline to compare with HypCluster is to run FedAvg $k$ times, and ensemble\footnote{\label{ensemble}There are many ways to combine multiple models, e.g., average the model outputs. Here we use the same procedure as in HypCluster, where each client selects the model with the lowest local loss, and use the selected model in inference.} the $k$ models. We also include this baseline in Table~\ref{table:cross-device}, and discuss it in Observation 3 below. 

\begin{figure}
    \centering
    \begin{subfigure}[b]{0.7\textwidth}
    \includegraphics[width=\textwidth]{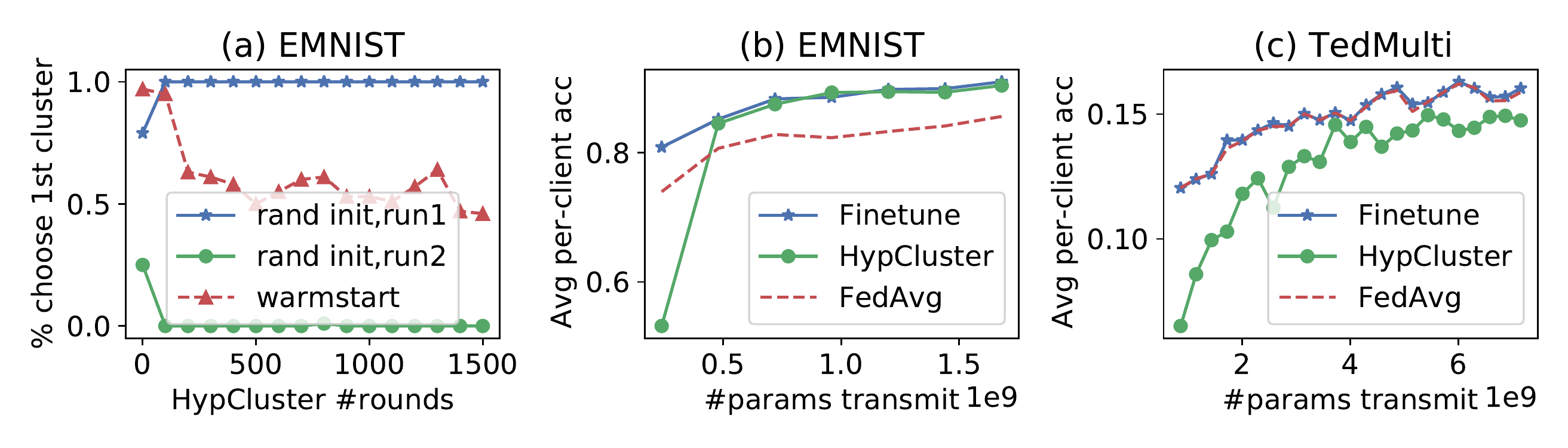}
    \end{subfigure}
    \hfill
    \begin{subfigure}[b]{0.28\textwidth}
    \includegraphics[width=\textwidth]{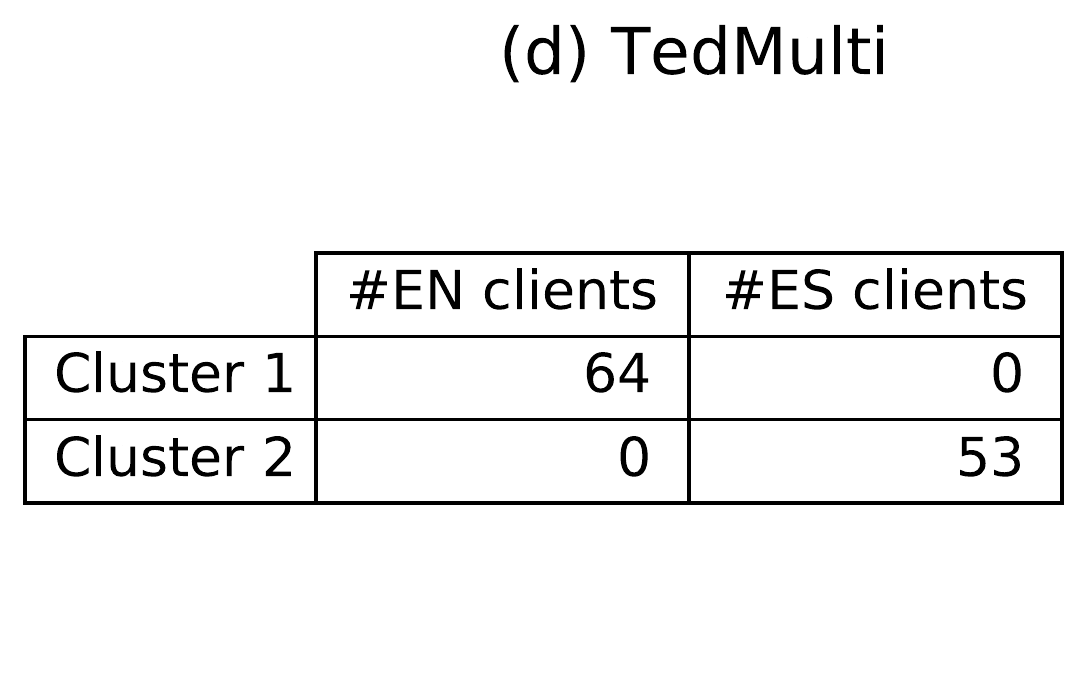}
    \end{subfigure}
    \caption{(a) Warmstart helps mitigate the mode collapse issue. (b--c) Plot of the average per-client accuracy as a function of communication cost for EMNIST and TedMulti. (d) Number of English and Spanish clients in the two clusters learned by HypCluster on TedMulti dataset.}
    \label{fig:hypcluster}
\end{figure}

\textbf{Observation 1: ``Mode collapse'' can hurt training.} In Figure~\ref{fig:hypcluster}(a), we monitor the percentage of clients that choose the first cluster when running HypCluster with random initialization on EMNIST. After a few rounds, all clients will always choose the same cluster. The same ``mode collapse'' issue happens to all the other datasets. One way to mitigate this issue (as shown in Figure~\ref{fig:hypcluster}(a)) is to warm start HypCluster with models trained by FedAvg. In our experiments, we train a few rounds of FedAvg for $k$ times, and use the $k$ models to warm start HypCluster (see Appendix~\ref{app:hparam} for the hyperparameters.). However, mode collapse can still happen even when warm start is used, especially for $k>2$. How to effectively train HypCluster without mode collapse in real-world cross-device settings remains an open problem.

\textbf{Observation 2: High communication cost per round.} A limitation of HypCluster is that in each round, the server needs to broadcast $k$ models (where $k$ is the number of learned clusters) to the clients, and hence, incurs $k$ times the communication cost of FedAvg\footnote{In real-world cross-device FL settings, reducing the communication cost between the server and mobile devices can help reduce the latency due to stragglers (see Section 5.1 in~\cite{wang2021field} for more discussions).}. In Figure~\ref{fig:hypcluster}(b--c), we plot the average per-client accuracies with respect to the total number of parameters communicated (including those used for training the warm start models) for EMNIST and TedMulti. While TedMulti eventually achieves a slightly better average accuracy than FedAvg (Table~\ref{table:benchmark-overview}), it is arguably worse than FedAvg from a accuracy-communication ratio perspective. While one can reduce communication costs via weight-sharing~\cite{ghosh2020efficient}, we found that this increased the likelihood of mode collapse\footnote{One possible reason is that during training, the shared layers may produce features specifically tied to one cluster. To avoid this correlation, one may want to learn invariant features~\cite{arjovsky2019invariant} that are good for all clusters.}. In future work it would be interesting to explore connections between mode collapse and weight sharing, and consider how to best train HypCluster in constrained networks.

\textbf{Observation 3: Difficulty in interpreting the learned clusters.} As mentioned in the beginning of Section~\ref{sec:cross-device-exp-cluster}, a natural baseline to compare with HypCluster is to run FedAvg $k$ times, and ensemble\footnoteref{ensemble} the $k$ models. Specifically, we train HypCluster with $k$ clusters for $y$ rounds, and compare it with training FedAvg $k$ times and each for $y$ rounds (note that both have the same communication cost). As shown in Table~\ref{table:cross-device}, the average per-client accuracy of ensembling $k$ FedAvg models is similar to that of HypCluster for 3/4 datasets, so a natural question is: does HypCluster actually learn a meaningful cluster structure of the underlying data? While it is generally hard to answer this question on an arbitrary dataset, we do know that TedMulti-EnEs has two natural clusters of clients, i.e., English and Spanish users. Figure~\ref{fig:hypcluster}(d) lists the number of English and Spanish clients (total 117 test clients) in the two learned clusters. HypCluster indeed uncovers the two underlying clusters on TedMulti-EnEs.

\textbf{Observation 4: Sensitive to distribution shift.} We follow the same ID and OOD evaluation procedure in Section~\ref{sec:cross-device-exp-FT}. As shown in Figure~\ref{fig:id_ood_tradeoff}, HypCluster seems to give a worse ID-OOD tradeoff (i.e., achieves a lower OOD accuracy at the same ID accuracy) compared to the other personalization algorithms.

\subsection{FedAvg + kNN-Per}\label{sec:cross-device-exp-knn}

We now discuss the results of kNN-Per, which are shown in Table~\ref{table:cross-device} as well as in Figure~\ref{fig:box-plot}.

\begin{figure}[ht]
    \centering
    \includegraphics[width=\linewidth]{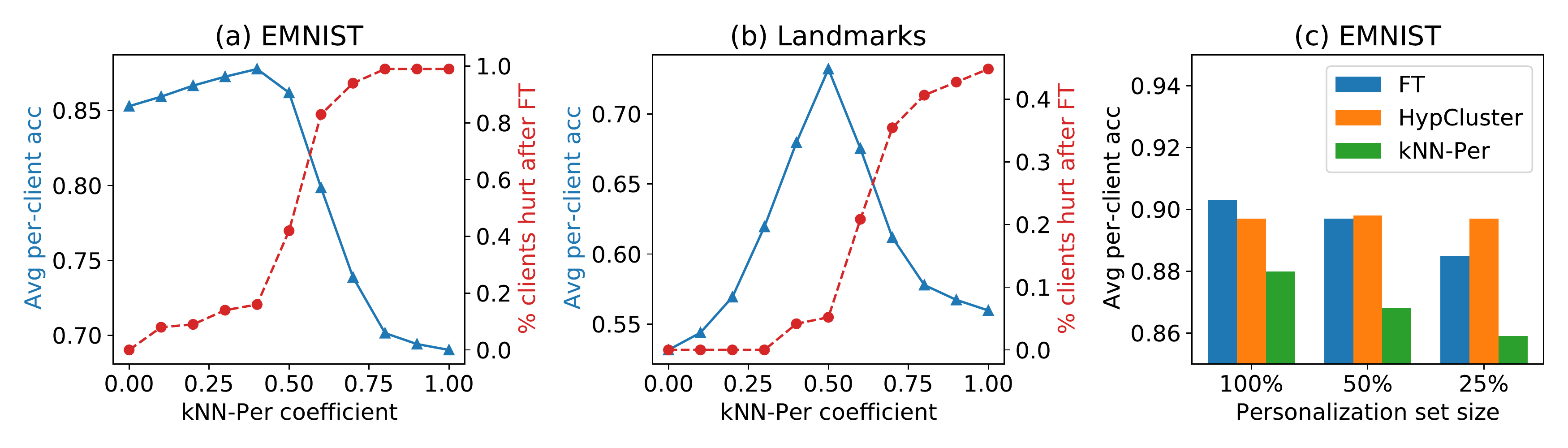}
    \caption{(a--b) Different metrics may be improved by different hyperpamater values: the "average per-client accuracy" metric usually prefers a moderately large interpolation coefficient while the "clients hurt" metric prefers a small coefficient. (c) The personalization accuracy drops for fine-tuning and kNN-Per when each client has fewer examples to personalize the model. HypCluster is robust to this change because it only needs the personalization set to choose the best model instead of training a new model.}
    \label{fig:knn_per_hparams}
\end{figure}

\textbf{Observation 1: Clients may be hurt.} In Table~\ref{table:cross-device}, besides the per-client accuracy, we also list the percentage of clients "hurt" after personalization (i.e., test clients whose accuracy on the local evaluation set\footnoteref{split} after kNN-Per is lower than that of FedAvg). As shown in Table~\ref{table:cross-device}, compared to fine-tuning, kNN-Per has a lower "per-client acc" but a higher "\% clients hurt" for 3/4 datasets. The two reasons why clients may hurt after fine-tuning given in Section~\ref{sec:cross-device-exp-FT} also applies to kNN-Per: 1) The per-client accuracy metric is noisy due to local data scarsity; 2) The hyperparameter is chosen globally but each client has heterogeneous local distribution. Similar to fine-tuning, so far we assume all test clients will perform kNN-Per. Designing a good set of conditions as to when a client should perform kNN-Per may help reduce the fraction of client hurt (see "Remark: Acceptance criteria and robust personalization" in Section~\ref{sec:cross-device-exp-FT}). 

\textbf{Observation 2: Difficulty in hyperparameter tuning.} Following~\cite{marfoq2022personalized}, we set $k$ (the number of nearest neighbors) to 10 for all kNN-Per experiments\footnote{\cite{marfoq2022personalized} shows that the performance of kNN-Per is pretty robust to the choice of $k$.}. Similar to fine-tuning\footnoteref{per-client-tune}, we tune the interpolation coefficient globally, i.e., all clients use the same coefficient\footnote{We also tried tuning a per-client specific coefficient but found this performed worse than global tuning due to limited amount of local data in the cross-device setting.}. The interpolation coefficient is a scalar in [0,1] (see Eq.(7) of~\cite{marfoq2022personalized}), where 0 means that no personalization and 1 means that using the local kNN model for inference. Figure~\ref{fig:knn_per_hparams}(a--b) plot the two metrics "average per-client accuracy" and "\% client hurt" for different interpolation coefficients, which shows that the two metrics may prefer different hyperparameter values. Section~\ref{sec:cross-device-exp-FT} points out two other reasons why hyperparameter tuning is difficult for fine-tuning. Both of them apply to kNN-Per as well, including the noisy local metrics and potentially extra hyperparameters.

\textbf{Observation 3: Sensitive to distribution shift.} We follow the same ID and OOD evaluation procedure in Section~\ref{sec:cross-device-exp-FT}. As shown in Figure~\ref{fig:id_ood_tradeoff}, kNN-Per seems to give a similar or better ID-OOD tradeoff (i.e., achieves a higher OOD accuracy at the same ID accuracy) compared to the other personalization algorithms. Nevertheless, it is still worth exploring methods that improve the OOD robustness during personalization, as discussed in the "Remark: Acceptance criteria and robust personalization" in Section~\ref{sec:cross-device-exp-FT}.

\textbf{Observation 4: Performance drops when clients have smaller personalization set\footnoteref{split}.} This has been discussed previously in Section~\ref{sec:cross-device-exp-FT} and illustrated in Figure~\ref{fig:knn_per_hparams}(c).

\section{Cross-Silo Experiments}\label{sec:cross-silo-exp}

We now consider baselines for the cross-silo personalization methods and datasets in 
\name. 

\begin{wrapfigure}[15]{r}{0.3\textwidth}
     \centering
     \includegraphics[width=\linewidth]{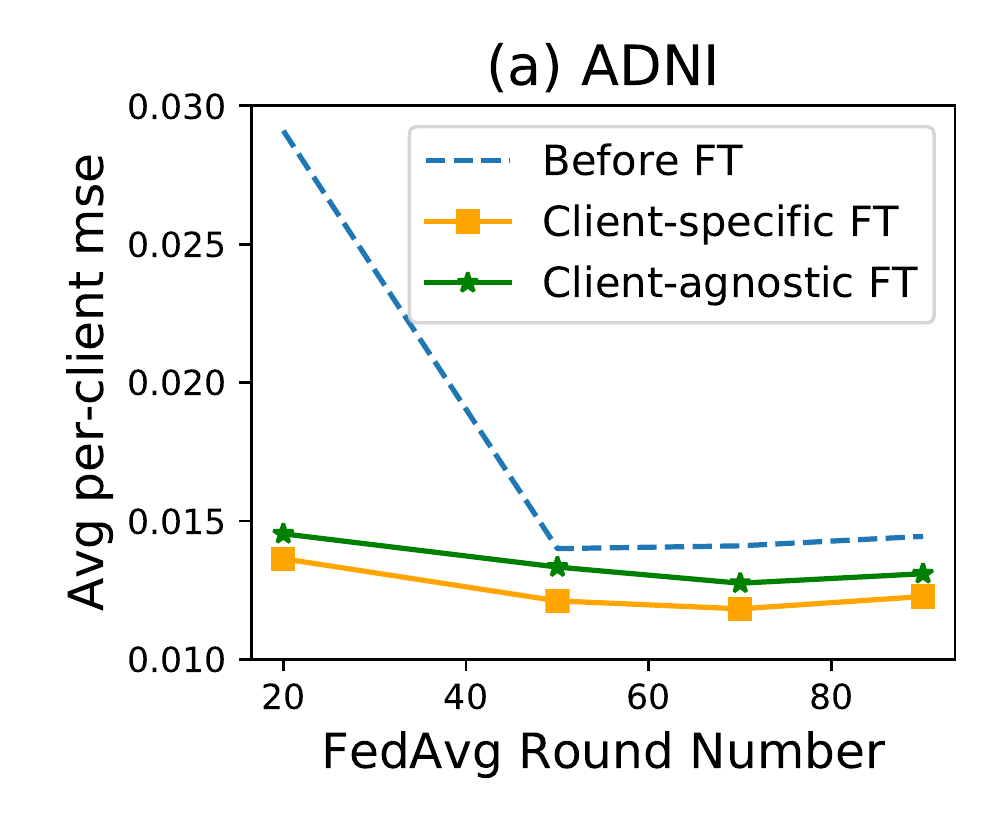}
     \caption{Tuning the fine-tuning hyperpameters in a per-client vs global manner.}
     \label{fig:cross-silo}
\end{wrapfigure}

\textbf{Training and evaluation setup.}
As mentioned in Section~\ref{sec:benchmark-data}, in the cross-silo setting, the same silos appear in both training and inference; to evaluate this setting, we split each client's local examples into three sets: train/validation/test. We train on the training sets across all clients, tune hyperparameters on the validation sets (see Appendix~\ref{app:hparam:silo}), and report metrics on the test sets. Unlike the cross-device setting, in the cross-silo setting, each client typically has sufficient compute power and a large number of local examples. This allows us to tune fine-tuning hyperparameters in a per-client manner (e.g., every client chooses their own hyperparameters based on their validation metric), as opposed to tuning globally\footnoteref{per-client-tune} in the cross-device setting mentioned in Section~\ref{sec:cross-device-exp}.

\begin{table}[t!]
  \renewcommand{\arraystretch}{1.1}
  \caption{Summary of experimental results on the cross-silo datasets. Similar to Table~\ref{table:cross-device}, we report the per-client test metric (mean $\pm$ standard deviation) \emph{across the clients}, so the standard deviation can be viewed as a fairness metric~\cite{li2021ditto, mansour2020three}. Each metric value is averaged over 5 independent runs with different random seeds (see Table~\ref{apptable:cross-silo} for standard deviations over the 5 runs). Vehicle uses accuracy as the metric while ADNI and School use mean squared error (MSE).  Appendix~\ref{app:hparam} has the tuned hyperparameters.
  } 
  \label{table:cross-silo}
  \centering
  \begin{tabular}{p{2.3cm}|p{4.8cm}p{2.4cm}p{2.4cm}p{2.4cm}}
    \toprule
    \textbf{Algorithm} & \textbf{Metrics}  & \textbf{Vehicle} (\textbf{\textcolor{OliveGreen}{acc}}) & \textbf{School} (\textbf{\textcolor{BrickRed}{mse}}) & \textbf{ADNI} (\textbf{\textcolor{BrickRed}{mse}}) \\
    \midrule
    Local training & Per-client metric & 0.9367$\pm$.0248  & 0.0121$\pm$.0059 & 0.0177$\pm$.0106 \\
    \midrule
    \multirow{7}{2.3cm}{FedAvg + Fine-tuning (FT)} & Per-client metric before FT & 0.8859$\pm$.0833  & 0.0130$\pm$.0068 & 0.0141$\pm$.0090  \\
    & Per-client metric after FT  &  0.9385$\pm$.0253  & 0.0116$\pm$.0056 & 0.0124$\pm$.0082 \\
    & \% clients "hurt" after FT  & 4.4\% &  33.0\% & 0  \\
    & FT all layers vs last layer & N/A & N/A  & no difference \\
    \cline{2-5}
    &\multicolumn{4}{c}{\textit{\textbf{Practical concerns:} Similar to those in the cross-device setting (Table~\ref{table:cross-device});}} \\
    &\multicolumn{4}{c}{\textit{Tuning per-client FT hyperparameter may outperform tuning globally (\Cref{fig:cross-silo}).}} \\
    \midrule
    \multirow{7}{2.3cm}{HypCluster / IFCA} & Per-client metric  & 0.9246$\pm$.0288  & 0.0112$\pm$.0053  &  0.0137$\pm$.0093  \\
    & No. tuned clusters ($k$) & 4 & 3  & 2 \\
    & \% clients largest cluster  & 49.6\% & 44.6\% &  78\%  \\
    & Warmstart from FedAvg & No & Yes & Yes \\
    & Per-client metric by ensembling $k$ FedAvg models  &  0.8851$\pm$.0828  &  0.0129$\pm$.0066  &  0.0133$\pm$.0091 \\
    \cline{2-5}
    &\multicolumn{4}{c}{\textit{\textbf{Practical concerns:} Similar to those in the cross-device setting (Table~\ref{table:cross-device}); on}}\\
    &\multicolumn{4}{c}{\textit{Vehicle and School, mode collapse may occur less potentially from using a linear model.} }\\
    \midrule
    \multirow{3}{2.3cm}{FedAvg + kNN-Per} & Per-client metric & 0.9228$\pm$.0287  & 0.01163$\pm$.0055 & 0.0126$\pm$.0096  \\
    & \% clients "hurt"  &  22.6\%  & 37.4\% & 37.8\% \\
    \cline{2-5}
    &\multicolumn{4}{c}{\textit{\textbf{Practical concerns:} Similar to those in the cross-device setting (Table~\ref{table:cross-device}).}} \\
    \midrule
    {MTL (Ditto)} & Per-client metric  & 0.9377$\pm$.0218 & 0.0114$\pm$.0053  &  0.0134$\pm$.0063   \\
    {MTL (Mocha)} & Per-client metric  &  0.9371$\pm$.0244 & 0.0121$\pm$.0059  & N/A    \\
    \bottomrule
  \end{tabular}
\end{table}

\textbf{Results.} 
Table~\ref{table:cross-silo} reports the cross-silo experimental results in the same format as in Table~\ref{table:cross-device}. Most of the practical concerns of FedAvg+Fine-tuning and HypCluster discussed in Section~\ref{sec:cross-device-exp} are still applicable here, e.g., HypCluster is still difficult to train due to the mode collapse issue (although it may happen less frequently when training simple linear models on Vehicle and School). In addition to the shared practical concerns, we observe four key trends specific to the cross-silo setting.

\textbf{Observation 1: Effectiveness of local training.} Local training may be a strong baseline in cross-silo settings. If we consider client(silo)-level differential privacy, it has an extra benefit of no privacy cost.

\textbf{Observation 2: Importance of personalization.} The four personalization algorithms (FedAvg + Fine-tuning, HypCluster, kNN-Per, and MTL) all achieve better mean accuracy (or MSE) than that of FedAvg. In terms of fairness\footnoteref{fairness}, the four personalization algorithms achieve better (i.e., smaller per-client metric standard deviation in Table~\ref{table:cross-silo}) or similar fairness metric than that of FedAvg.

\textbf{Observation 3: Performance of MTL.} We see that the (stateful) MTL methods could yield competitive performance with other stateless personalization algorithms on the three cross-silo datasets, with potentially less hyperparameter tuning.

\textbf{Observation 4: Effectiveness of client-specific fine-tuning.} For FedAvg+Fine-tuning, we compared tuning the fine-tuning hyperparameters (i.e., fine-tuning learning rate and the number of fine-tuning epochs) in a per-client (every client chooses their own hyperparameter) vs global (all clients share the same hyperparameter) manner. Using client-specific fine-tuning hyperparameters can be better than tuning globally when each client's local data is large (see the ADNI result in Figure~\ref{fig:cross-silo}). 
\section{Discussion and Open Directions}\label{sec:conclusion}

In this work we present \name, a large-scale benchmark for personalized FL covering both cross-device and cross-silo settings. \name provides a reproducible, end-to-end experimental pipeline including data preprocessing, algorithms, evaluation metrics, and tuned hyperparameters\footnoteref{code}. Beyond these baselines, our experiments highlight key insights about personalized FL (see the detailed summary in Appendix~\ref{app:conclusion}). These insights suggest several directions of future work, such as: 

\begin{itemize}
\item The notion of the "best" method (or the "best" hyperparameter of the same method as shown in Figure~\ref{fig:tune_ft_hparams}) can change depending on the evaluation metric or setting (see Table~\ref{table:cross-device} and Table~\ref{table:cross-silo}). In this paper we report several different evaluation metrics, including the average per-client accuracy, a fairness\footnoteref{fairness} metric, the fraction of clients hurt by personalization, the communication cost (Figure~\ref{fig:hypcluster}), robustness to distribution shift (Figure~\ref{fig:id_ood_tradeoff}), and robustness to the number of local examples (Figure~\ref{fig:knn_per_hparams}(c)). A critical direction is thus to develop systematic evaluation schemes for personalized FL (i.e., mean accuracy alone is not enough). 

\item Existing literature often overlook or obfuscate the practical complexities of deploying personalized FL algorithms in real-world settings (see the practical concerns of each method summarized in Table~\ref{table:cross-device} and Table~\ref{table:cross-silo}). Designing new practical personalized FL methods that take these considerations into account is an important direction to democratize FL. 

\item Developing techniques to train and interpret clustering methods such as HypCluster without mode collapse (Section~\ref{sec:cross-device-exp-cluster}) is a necessary step to make these approaches more effective in practice. 

\item Tradeoffs exists between adapting a client's personalized model to the current local distribution and generalizing to future distributions ( Figure~\ref{fig:id_ood_tradeoff}), which is worth exploring in greater detail (see the remark on "acceptance criteria and robust personalization" in Section~\ref{sec:cross-device-exp-FT}).

\item Given the observed benefits of  per-client hyperparameter tuning in cross-silo FL (Figure~\ref{fig:cross-silo}), it may be beneficial to develop similar, scalable approaches for hyperparameter tuning in cross-device FL.
\end{itemize}

Finally, we note that the area of benchmarking itself can be improved in future iterations. For example, we hope that \name can inspire benchmarking of additional evaluation metrics such as privacy, other notions of fairness\footnoteref{fairness}, and robustness, and additional datasets and applications.

\section*{Acknowledgements}
We are grateful to Zachary Garrett, Jakub Kone{\v{c}}n{\`y}, H. Brendan Mcmahan, Sewoong Oh, Daniel Ramage, Keith Rush, and Ananda Theertha Suresh for helpful discussions and comments.
\bibliography{ref.bib}
\bibliographystyle{plainnat}
\newpage
\appendix
\section{Background on Personalized Federated Learning}\label{app:background}
Traditionally, federated learning objectives consider fitting a single global model, $w$, across all local data in the network. The aim is to solve:
\begin{equation}\label{obj:global}
    \min_w \, G(F_1(w), \dots\, F_K(w)) \, , 
\end{equation}
where $F_k(w)$ is the local objective for client $k$, and  $G(\cdot)$ is a function that aggregates the local objectives $\{F_k(w)\}_{k \in [K]}$ from each client. For example, in FedAvg~\cite{mcmahan2017communication}, $G(\cdot)$ is typically set to be a weighted average of local losses, i.e., $\sum_{k=1}^K p_k F_k(w)$, where $p_k$ is a pre-defined non-negative weight such that $\sum_k p_k=1$. However, in general, each client may generate data $x_k$ via a distinct distribution $\mathcal{D}_k$, i.e., $F_k(w) := \mathbb{E}_{x_k \sim \mathcal{D}_k}\left[f_k(w; x_k)\right]$. To better account for this heterogeneity, it is therefore increasingly common to consider techniques (described below) that learn personalized, client-specific models, $\{w_k\}_{k \in [K]}$ across the network. 

A distinguishing factor of personalized FL methods is whether the approach requires any variables to be maintained on participating clients from one round to another~\cite{wang2021field}. As discussed in Section~\ref{sec:background}, whereas stateless approaches may be applicable to either cross-device or cross-silo FL, stateful approaches are more appropriate for cross-silo settings given the typical size and configuration of the network. Besides stateful vs stateless, the personalized FL algorithms can be categorized into model-agnostic and model-specific. Model-specific approaches target a specific model and usually require domain-specific information~\cite{singhal2021federated, zhu2021diurnal, jain2021differentially}. The current version of \name focuses on benchmarking personalization algorithms that can work without assumptions on the model or application scenario. We discuss major model-agnostic approaches in stateful vs. stateless personalized FL below, and defer readers to the recent surveys~\cite{tan2022towards}\cite[\S7.5]{wang2021field} for more related work on personalized FL.

\textbf{Stateful Approaches.} A common class of stateful personalized FL methods are \textit{multi-task learning (MTL)} methods. These methods view each client (or group of clients) as a separate `task', and aim to jointly learn task-specific models while exploiting similarities/differences between tasks. The idea of solving multiple  learning tasks simultaneously was first popularized by~\citet{Caruana:1998ml} in the 90's, who described multi-task learning as a technique in which related tasks act as a form of inductive bias to improve generalization. Many approaches can be captured in the following general and widely-used formulation, known as multi-task relationship learning~\cite{zhang2017survey,Zhang:2010ac,smith2017federated}:
\begin{equation}
 \label{eq:obj}
 \min_{\bm{W}, \bm{\Omega}} \, \left\{ \sum_{k=1}^K \sum_{i=1}^{n_k} \ell_k(\bm{w}_k; \bm{x}_k^i, y_k^i) +  \mathcal{R}(\bm{W}, \bm{\Omega}) \right\} \, .
 \end{equation}
 Here $\bm{W} := [\bm{w}_1, \dots, \bm{w}_K] \in\mathbb{R}^{d\times K}$ is a matrix whose $k$-th column is the weight vector for the $k$-th task (in this case, the model on client $k$). The matrix $\bm{\Omega} \in \mathbb{R}^{K\times K}$ models relationships amongst tasks, and is either known a priori or estimated while simultaneously learning task models. MTL problems differ based on their
assumptions on $\mathcal{R}$, which takes $\bm{\Omega}$ as input and promotes some suitable structure amongst the tasks. In federated learning, a number of specific MTL instantiations have been proposed, including variants of general relationship learning~\cite{smith2017federated,huang2021personalized}, cluster-regularized MTL~\cite{smith2017federated,sattler2020clustered}, global-regularized MTL~\cite{li2021ditto,yu2020salvaging}, and mean-regularized MTL~\cite{hanzely2020federated,hanzely2020lower,dinh2020personalized}.

Beyond these approaches, another common variant of MTL, particularly for deep learning problems, is `hard parameter' sharing approaches~\cite{ruder2017overview}. These methods consider splitting the model architecture itself into two components:  a shared part that is jointly learned by all
clients, and a local part that is personalized to each client. The local portion can be simply fine-tuned (as discussed below), or may be learned in conjunction with the shared portion by saving the local state on each client at every round, in which case they would also fall under the category of stateful FL~\cite{liang2020think,arivazhagan2019federated,hanzely2021personalized}. 


\textbf{Stateless Approaches.} In the stateless category, one of the most common forms of personalization is simple fine-tuning. With fine-tuning, a shared model is trained and deployed on each client, and the model is then fine-tuned or adapted locally to the client's data. In its simplest form the deployed model could be a global model trained via a typical procedure such as FedAvg~\cite{mcmahan2017communication, reddi2021adaptive}, and additional iterations of a stochastic optimizer such as mini-batch SGD can be run locally after deployment. However, it is also natural to consider \textit{meta-learning} approaches as part of this workflow~\cite{jiang2019improving,fallah2020personalized,khodak2019adaptive,charles2021convergence}, which are specifically designed to learn an algorithm that can solve a new task with a small number of training samples. These approaches can be particularly useful in cross-device FL settings, where each client may generate only a small number of training points, and it may be necessary to deploy and adapt models to clients that did not participate in training.

Finally, unlike the stateful MTL-based clustering approaches discussed above, it is also possible to consider clustering variants that don't require state to be maintained. In particular, a common approach is to maintain multiple global models and to have participating clients determine which of the models is best suited to their local data, thus forming a natural clustering amongst the clients~\cite{mansour2020three, ghosh2020efficient}.


\section{Datasets}\label{app:data}

\begin{figure}[ht]
    \centering
    \includegraphics[width=\linewidth]{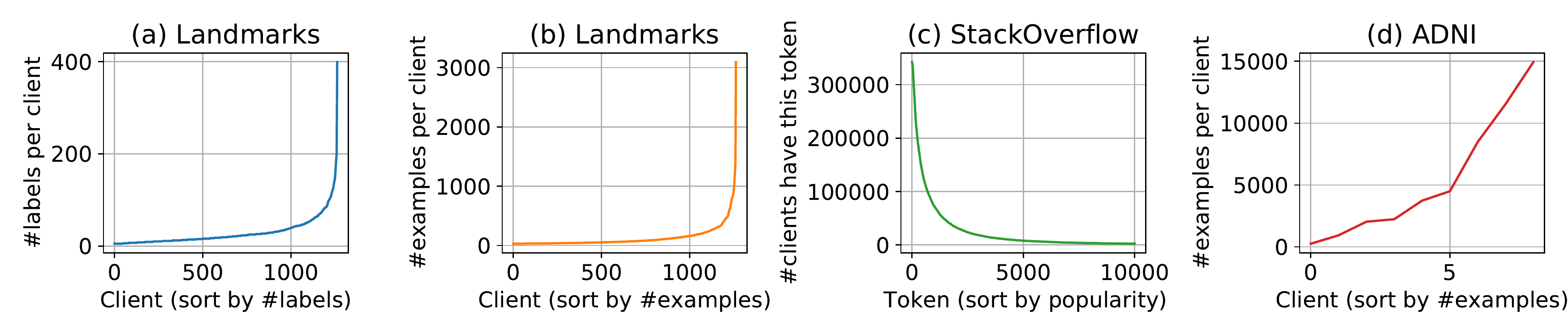}
    \caption{The chosen federated datasets (Table~\ref{table:benchmark-overview}) have heterogeneous local distributions.}
    \label{appfig:data-hetero}
\end{figure}

All the datasets used in \name are listed in Table~\ref{table:benchmark-overview}. They do not contain personally identifiable information or offensive content. They all have natural per-client data partitions, which will be discussed in more details below. Figure~\ref{appfig:data-hetero} shows that each client's local data distribution is distinct from each other, a common property of real-world federated datasets (see the discussions on "Heterogeneity" in Section~\ref{sec:background} and Appendix~\ref{app:background}).

\subsection{Cross-device}

We describe the cross-device datasets in Table~\ref{table:benchmark-overview}. We will also mention how we split the clients into train, validation, and test clients (a key preprocessing step described in Section~\ref{sec:benchmark-data}).
\begin{itemize}[leftmargin=*]
\item \textbf{EMNIST} is first processed by the LEAF benchmark~\cite{caldas2018leaf}. We use the version from~\cite{emnist}, which contains 3400 writers/clients and a total of 671,585 images\footnote{Note that in the EMNIST dataset provided by~\cite{emnist}, each client's examples are already split into a training set and a small test set, for simplicity, we only use the training set of each client in our experiments.}. Each client has on average 200 images (28-by-28 pixels) of hand-written digits and characters (62 classes). We use a CNN model~\cite{lecun1989backpropagation} with 1M parameters (same as~\cite{charles2021large}). The 3400 clients are randomly split into 2500/400/500 as train/validation/test.

\item \textbf{StackOverflow} is a large-scale federated language dataset. Each client is a user in the Stack Overflow online forum. The examples are the questions and answers posted by this user. Each example is a sentence. The task is next word prediction. We use an LSTM model~\cite{hochreiter1997long} with 4M parameters and 10k vocabulary size (same as~\cite{charles2021large}). The version from~\cite{stackoverflow} contains 342k train clients and 38k held-out clients. We randomly split the held-out clients into 10k/28k for validation/test.

\item \textbf{Landmarks} is a federated image classification dataset processed by~\citet{hsu2020federated} (from the 2019 Landmark Recognition Challenge~\cite{weyand2020google}). Each client is a Wikipedia contributor. The images are the landmark (e.g., famous monuments and mountains) photos uploaded by the photographers. The per-user data distribution naturally varies based on the photographer's geographic location. The dataset contains 164k images, 2028 landmark labels, and 1262 clients. We use a MobileNetV2 model~\cite{sandler2018mobilenetv2} with 4M parameters (same as~\cite{hsu2020federated}). We randomly split the clients into 1112/50/100 as train/validation/test. 

\item \textbf{TedMulti-EnEs} is a subset of the TedMultiTranslate dataset. The original TedMultiTranslate is a multilingual (60 languages) dataset derived from the TED Talk transrcipts~\cite{qi2018and}. We only use two languages English and Spanish (and hence, "EnEs" in the name). This dataset has not been used in any federated learning experiment before, so we need to do more preprocessing work. We first partition the clients by the TED Talk author and language (a client contains either English data or Spanish data but not both\footnote{\label{app:tedmulti}The processed dataset has two clusters of clients: English/Spanish. See Section~\ref{sec:cross-device-exp-cluster} for more discussions.}). Each example is a sentence from the talk. The task is next word prediction. We build a vocabulary of size 15k containing both English and Spanish words (these words appear in at least 20 clients). We use a transformer model~\cite{vaswani2017attention} with 4M parameters. Clients with less than 20 examples are removed. We randomly split the 4184 clients into 3969/98/117 as train/validation/test.
\end{itemize}

\subsection{Cross-silo}

\begin{itemize}[leftmargin=*]
\item \textbf{Vehicle} dataset, originally collected by~\citet{duarte2004vehicle}, is a binary classification dataset containing measurements of road segments from a distributed network of 23 vehicle sensors with a total of 43,695 feature vectors for classifying the type of the passing vehicles. Following~\citet{smith2017federated}, we treat each sensor as a task (data silo) and use 50 acoustic and 50 seismic features as inputs to a linear SVM model. 
It is suitable for cross-silo FL because of the small number of silos and sufficiently large local datasets for local training. We split each client(silo)'s local dataset into 70\%/15\%/15\% train/validation/test sets.

\item \textbf{School} dataset, originally collected by the now-defunct Inner London Education Authority,\footnote{\url{https://en.wikipedia.org/wiki/Inner_London_Education_Authority}}
is a regression dataset for predicting the exam scores of 15,362 students distributed across 139 secondary schools.
Each school has records up to 251 students with each student described by a 28-dimensional feature vector capturing attributions such as the school ranking, student birth year, and whether the school provided free meals. Simple linear regression models suffice for reasonable performance. We split each client(silo)'s local dataset into 70\%/15\%/15\% train/validation/test sets.

\item \textbf{ADNI} (Alzheimer's Disease Neuroimaging Initiative)\footnote{All research on the ADNI dataset was conducted by Tian Li. No other authors had access to this data. ADNI data were obtained from the Alzheimer’s Disease
Neuroimaging Initiative (ADNI) database (adni.loni.usc.edu). As such, the investigators
within the ADNI contributed to the design and implementation of ADNI and/or provided data
but did not participate in analysis or writing of this paper. A complete listing of ADNI
investigators can be found at:
http://adni.loni.usc.edu/wp-content/uploads/how\_to\_apply/ADNI\_Acknowledgement\_List.pdf. ADNI project was funded by National Institutes of Health Grant U01 AG024904 and
DOD ADNI (Department of Defense award number W81XWH-12-2-0012). } is a public medical dataset containing various formats of data to help advance the study of Alzheimer’s disease.\footnote{\href{https://adni.loni.usc.edu/}{https://adni.loni.usc.edu/}} The task is to predict Standardized Uptake Value Ratio (SUVR) from PET scans of human brains.
We treat each scanner vendor as a silo, and there are 9 silos in total. We specifically use a subset of PET scans (with AV45 and preprocessing step \texttt{`Coreg, Avg, Std Img and Vox Siz, Uniform Resolution'}) that have existing labels obtained from UC Berkeley study in the database. We  convert the data into .png format, normalize each pixel value to $(0,1)$, and rescale each image into size 32$\times$32.  We split each client(silo)'s local dataset into 80\%/10\%/10\% train/validation/test sets.
\end{itemize}

\subsection{License/Usage Information}
\begin{itemize}[leftmargin=*]
\item \textbf{EMNIST.} The Extended MNIST dataset\footnote{\url{https://www.westernsydney.edu.au/icns/reproducible_research/publication_support_materials/emnist}} is a variant of the original NIST Special Database 19\footnote{\url{https://www.nist.gov/srd/nist-special-database-19}}. In this paper, we use a federated version of this Extended MNIST dataset, which is processed and provided by the LEAF\footnote{\url{https://github.com/TalwalkarLab/leaf}} benchmark~\cite{caldas2018leaf} under the BSD-2-Clause license. 

\item\textbf{StackOverflow.} This dataset is derived from the Stack Overflow Data hosted by Kaggle\footnote{\url{ https://www.kaggle.com/stackoverflow/stackoverflow}}, under the CC BY-SA 3.0 license. In this paper, we use the federated version of this dataset processed by Tensorflow Federated~\cite{stackoverflow}, which is open-sourced under the Apache-2.0 license.

\item\textbf{Landmarks.} The original Google Landmarks Dataset v2~\cite{weyand2020google} can be accessed from GitHub\footnote{\url{https://github.com/cvdfoundation/google-landmark}}: "The annotations are licensed by Google under CC BY 4.0 license. The images listed in this dataset are publicly available on the web, and may have different licenses". In this paper, we use a subset of the full dataset processed by~\citet{hsu2020federated}: the data split function is available on GitHub\footnote{\url{https://github.com/google-research/google-research/tree/master/federated_vision_datasets}}  under the Apache-2.0 license.

\item\textbf{TedMulti-EnEs.} The original TedMultiTranslate dataset is derived from the TED Talk transcripts~\cite{qi2018and}. The TED Talk usage policy can be found from their website\footnote{\url{https://www.ted.com/about/our-organization/our-policies-terms/ted-talks-usage-policy}}. The TedMultiTranslate dataset can be downloaded from GitHub\footnote{\url{https://github.com/neulab/word-embeddings-for-nmt}}. We were unable to find the license information. In this paper, we extract a subset of this dataset and group the examples by talk author and language\footnoteref{app:tedmulti}. This preprocessing pipeline is available from our benchmark code\footnote{\codeurl} under the Apache-2.0 license.

\item\textbf{Vehicle.}\enspace
The Vehicle dataset is made publicly available by the original authors as a research dataset~\cite{duarte2004vehicle}. It has been subsequently used in recent work (e.g.~\cite{smith2017federated}). License information was unavailable online. A copy of the dataset may be obtained from these URLs.\footnote{\url{https://web.archive.org/web/20110515133717/http://www.ece.wisc.edu:80/~sensit/}}\textsuperscript{,}\footnote{\url{https://web.archive.org/web/20200128092656/http://www.ecs.umass.edu:80/~mduarte/Software.html}}

\item\textbf{School.}\enspace
The School dataset was collected by a now-defunct entity and we were unable to find license information online. 
The dataset has been made freely available online and can be obtained from this URL.\footnote{\url{https://web.archive.org/web/20060718012309/http://www.mlwin.com/intro/datasets.html}} Alternatively, a copy may also be obtained from~\cite{zhou2011malsar} under the GNU GPL v2 license.

\item\textbf{ADNI.}\enspace
The term of use of ADNI can be found on their website.\footnote{\href{https://adni.loni.usc.edu/terms-of-use/}{https://adni.loni.usc.edu/terms-of-use/}}
\end{itemize}

\section{Hyperparameters and Implementation Details} \label{app:hparam}


In this section, we describe the computation resources, the hyperparameter grids and the best hyperparameters for all the experiments described in Section~\ref{sec:cross-device-exp} and Section~\ref{sec:cross-silo-exp}.

\subsection{Cross-device} \label{app:hparam:device}
The definitions of all hyperparameters can be found in \verb|finetuning_trainer.py| and \verb|hypcluster_trainer.py| under \url{https://github.com/google-research/federated/tree/master/personalization_benchmark/cross_device}.

\textbf{Common hyparameters.} The following hyperparameters are fixed across all cross-device experiments. Note that we
focus on \emph{FedAdam} here, i.e., a generalized version of FedAvg~\cite{reddi2021adaptive}, where the server
optimizer is Adam optimizer~\cite{kingma2014adam}, because \cite{charles2021large} shows FedAdam gives
good performance across different datasets.  
\begin{itemize}
    \item \verb|client_optimizer=[`sgd']|
    \item \verb|server_optimizer=[`adam']|
    \item \verb|server_adam_beta_1=[0.9]|
    \item \verb|server_adam_beta_2=[0.99]|
    \item \verb|train_epochs=[1]| \\ (This is the number of local training epochs performed by a client during a round of training.)
\end{itemize}

\textbf{Computation resources.} We summarize the computation resources allocated to run one experiment
(i.e., one point in the the hyperparameter grids) on each dataset: EMNIST (80 CPU cores); StackOverflow (400 GPU cores); Landmarks (16 GPUs); TedMulti-EnEs (2 GPUs). Note that the actual usage may be smaller than the allocated resources.

\subsubsection{FedAvg (i.e., FedAdam in our case) + Fine-tuning}\label{app:fedavg_finetuning_hparams}

The definitions of all the hyperparaemeters for running this algorithm can be found
in \verb|finetuning_trainer.py|. Since FedAvg + Fine-tuning is a two step process,
the hyperparameters contain FedAvg (i.e., FedAdam in our case) hyperparameters and fine-tuning hyperparameters.

For EMNIST and StackOverflow, we use the best FedAdam hyperparameters from~\cite{charles2021large}. For Landmarks,
we use the best FedAdam hyperparameters from~\cite{wang2021field}. For TedMulti-EnEs, we
tune the FedAdam hyperparameters from scratch.

We have four fine-tuning hyperparameters. Their names are all started with
\verb|finetune_|, including the optimizer used to fine-tune the model (where we focus
on SGD\footnoteref{ft-sgd}), the fine-tuning learning rate, and whether to only fine-tune the last
layer. We also need to tune the number of local epochs used to fine-tune the model - this value is automatically found by \verb|finetuning_trainer.py| by
postprocessing the validation metrics. Specifically, we compute the average
validation accuracy of the fine-tuned models after every fine-tuning epoch
(until \verb|finetune_max_epochs|), and then find the best fine-tuning epoch that
gives the highest average validation accuracy (see, e.g., Figure~\ref{fig:tune_ft_hparams}(a)\footnote{This figure shows how average test accuracy changes with respect to the fine-tuning epochs. Average validation accuracy follows a similar trend as the test accuracy.}). The best fine-tuned epoch will be
in the range \verb|[0, finetune_max_epochs]|, so all we need is to set a proper value
for \verb|finetune_max_epochs|.

\paragraph{EMNIST}\mbox{}\\

Fixed hyperparameters (we use the best FedAdam hyperparameters from~\cite{charles2021large}):
\begin{itemize}
    \item \verb|client_learning_rate=[0.1]|
    \item \verb|server_learning_rate=[0.001]|
    \item \verb|server_adam_epsilon=[0.001]|
    \item \verb|clients_per_train_round=[50]|
    \item \verb|train_batch_size=[20]|
    \item \verb|total_rounds=[1500]|
    \item \verb|valid_clients_per_round=[100]|
    \item \verb|test_clients_per_round=[100]|
    \item \verb|rounds_per_evaluation=[100]|
    \item \verb|rounds_per_checkpoint=[100]|
    \item \verb|finetune_optimzier=[‘sgd’]|
    \item \verb|finetune_max_epochs=[20]|
\end{itemize}

Tuned hyperparameters (best values are highlighted in \verb|**value**|):
\begin{itemize}
    \item \verb|finetune_learning_rate=[0.001, 0.003, **0.005**, 0.01, 0.05] |
    \item \verb|finetune_last_layer=[True, **False**]|
\end{itemize}

\paragraph{StackOverflow}\mbox{}\\

Fixed hyperparameters (we use the best FedAdam hyperparameters from~\cite{charles2021large}):
\begin{itemize}
    \item \verb|client_learning_rate=[1.0]|
    \item \verb|server_learning_rate=[0.1]|
    \item \verb|server_adam_epsilon=[0.001]|
    \item \verb|clients_per_train_round=[200]|
    \item \verb|train_batch_size=[16]|
    \item \verb|total_rounds=[1500]|
    \item \verb|valid_clients_per_round=[200]|
    \item \verb|test_clients_per_round=[200]|
    \item \verb|rounds_per_evaluation=[100]|
    \item \verb|rounds_per_checkpoint=[100]|
    \item \verb|finetune_optimzier=[‘sgd’]|
    \item \verb|finetune_max_epochs=[20]|
\end{itemize}

Tuned hyperparameters (best values are highlighted in \verb|**value**|):
\begin{itemize}
    \item \verb|finetune_learning_rate=[**10^(-1.0)**, 10^(-0.6), 10^(-0.2), 10^(0.2), | \\
    \verb|10^(0.6), 10^(1.0)]|
    \item \verb|finetune_last_layer=[True, **False**]|
\end{itemize}

\paragraph{Landmarks}\mbox{}\\

Fixed hyperparameters (we use the best FedAdam hyperparameters from~\cite{wang2021field}):
\begin{itemize}
    \item \verb|client_learning_rate=[0.01]|
    \item \verb|server_learning_rate=[10^(-2.5)]|
    \item \verb|server_adam_epsilon=[10^(-5)]|
    \item \verb|clients_per_train_round=[64]|
    \item \verb|train_batch_size=[16]|
    \item \verb|total_rounds=[30000]|
    \item \verb|valid_clients_per_round=[32]|
    \item \verb|test_clients_per_round=[96]|
    \item \verb|rounds_per_evaluation=[1000]|
    \item \verb|rounds_per_checkpoint=[1000]|
    \item \verb|finetune_optimzier=[‘sgd’]|
    \item \verb|finetune_max_epochs=[10]|
\end{itemize}

Tuned hyperparameters (best values are highlighted in \verb|**value**|):
\begin{itemize}
    \item \verb|finetune_learning_rate=[0.0001, 0.001, 0.005, **0.007**, 0.01, 0.03, 0.05]|
    \item \verb|finetune_last_layer=[True, **False**]|
\end{itemize}

\paragraph{TedMulti-EnEs}\mbox{}\\

Fixed hyperparameters:
\begin{itemize}
    \item \verb|clients_per_train_round=[32]|
    \item \verb|train_batch_size=[16]|
    \item \verb|total_rounds=[1500]|
    \item \verb|valid_clients_per_round=[98]|
    \item \verb|test_clients_per_round=[117]|
    \item \verb|rounds_per_evaluation=[30]|
    \item \verb|rounds_per_checkpoint=[50]|
    \item \verb|finetune_optimzier=[‘sgd’]|
    \item \verb|finetune_max_epochs=[20]|
\end{itemize}

Tuned hyperparameters (best values are highlighted in \verb|**value**|):
\begin{itemize}
    \item \verb|client_learning_rate=[10^(-2.5), 10^(-2), 10^(-1.5), **10^(-1)**, 10^(-0.5)]|
    \item \verb|server_learning_rate=[10^(-2.5), **10^(-2)**, 10^(-1.5), 10^(-1), 10^(-0.5)]|
    \item \verb|finetune_learning_rate=[**0.0005**, 0.0007, 0.001, 0.002, 0.003]|
    \item \verb|finetune_last_layer=[True, **False**]|
\end{itemize}

\subsubsection{HypCluster}
Definitions of all the hyperparameters for this algorithm can be found in \verb|hypcluster_trainer.py|. Because HypCluster with random initialization usually ends up with all clients choosing the same model (i.e., the mode
collapse issue shown in Figure~\ref{fig:hypcluster}), we will use models learned by FedAvg to warmstart HypCluster. Specifically, we will run FedAvg (with the hyperparameters in Appendix~\ref{app:fedavg_finetuning_hparams} above) for \verb|num_warmstart_fedavg_rounds|; repeat this for \verb|num_clusters| times, and use the models to warmstart HypCluster.

\paragraph{EMNIST}\mbox{}\\

Fixed hyperparameters:
\begin{itemize}
    \item \verb|clients_per_train_round=[50]|
    \item \verb|train_batch_size=[20]|
    \item \verb|total_rounds=[1500]|
    \item \verb|valid_clients_per_round=[100]|
    \item \verb|test_clients_per_round=[100]|
    \item \verb|rounds_per_evaluation=[100]|
    \item \verb|rounds_per_checkpoint=[100]|
    \item \verb|num_warmstart_fedavg_rounds=[100]|
\end{itemize}

Tuned hyperparameters (best values are highlighted in \verb|**value**|):
\begin{itemize}
    \item \verb|client_learning_rate=[0.01, 0.05, **0.1**, 0.2]|
    \item \verb|server_learning_rate=[0.0001, 0.0005, **0.001**, 0.002]|
    \item \verb|server_adam_epsilon=[**0.0001**, 0.0005, 0.001, 0.002]|
    \item \verb|num_clusters=[**2**, 3, 4, 5]|
\end{itemize}

\paragraph{StackOverflow}\mbox{}\\

Fixed hyperparameters:
\begin{itemize}
    \item \verb|clients_per_train_round=[200]|
    \item \verb|train_batch_size=[16]|
    \item \verb|total_rounds=[1500]|
    \item \verb|valid_clients_per_round=[200]|
    \item \verb|test_clients_per_round=[200]|
    \item \verb|rounds_per_evaluation=[100]|
    \item \verb|rounds_per_checkpoint=[100]|
    \item \verb|num_warmstart_fedavg_rounds=[100]|
\end{itemize}

Tuned hyperparameters (best values are highlighted in \verb|**value**|):
\begin{itemize}
    \item \verb|client_learning_rate=[0.1, **0.5**, 1.0, 2.0]|
    \item \verb|server_learning_rate=[**0.01**, 0.05, 0.1, 0.2]|
    \item \verb|server_adam_epsilon=[10^(-5), **10^(-4)**, 10^(-3), 10^(-2)]|
    \item \verb|num_clusters=[**2**, 3, 4, 5]|
\end{itemize}

\paragraph{Landmarks}\mbox{}\\

Fixed hyperparameters:
\begin{itemize}
    \item \verb|clients_per_train_round=[64]|
    \item \verb|train_batch_size=[16]|
    \item \verb|total_rounds=[30000]|
    \item \verb|valid_clients_per_round=[32]|
    \item \verb|test_clients_per_round=[96]|
    \item \verb|rounds_per_evaluation=[1000]|
    \item \verb|rounds_per_checkpoint=[1000]|
    \item \verb|num_warmstart_fedavg_rounds=[8000]|
\end{itemize}

Tuned hyperparameters (best values are highlighted in \verb|**value**|):
\begin{itemize}
    \item \verb|client_learning_rate=[10^(-3), **10^(-2.5)**, 10^(-2), 10^(-1.5)]|
    \item \verb|server_learning_rate=[10^(-3.5), **10^(-3)**, 10^(-2.5), 10^(-2)]|
    \item \verb|server_adam_epsilon=[10^(-6), 10^(-5), **10^(-4)**, 10^(-3)]|
    \item \verb|num_clusters=[**2**, 3, 4]|
\end{itemize}

\paragraph{TedMulti-EnEs}\mbox{}\\

Fixed hyperparameters:
\begin{itemize}
    \item \verb|clients_per_train_round=[32]|
    \item \verb|train_batch_size=[16]|
    \item \verb|total_rounds=[1500]|
    \item \verb|valid_clients_per_round=[98]|
    \item \verb|test_clients_per_round=[117]|
    \item \verb|rounds_per_evaluation=[30]|
    \item \verb|rounds_per_checkpoint=[50]|
    \item \verb|num_clusters=[2]|
    \item \verb|num_warmstart_fedavg_rounds=[100]|
\end{itemize}

Tuned hyperparameters (best values are highlighted in \verb|**value**|):
\begin{itemize}
    \item \verb|client_learning_rate=[10^(-2.5), 10^(-2), 10^(-1.5), **10^(-1)**, 10^(-0.5)]|
    \item \verb|server_learning_rate=[10^(-2.5), **10^(-2)**, 10^(-1.5), 10^(-1), 10^(-0.5)]|
    \item \verb|server_adam_epsilon=[**0.001**, 0.00001]|
\end{itemize}

\subsection{FedAvg + kNN-Per}

We use the model trained by FedAvg (same as "FedAvg + Fine-tuning") as the global model for personalization. Following~\cite{marfoq2022personalized}, we set the number of nearest neighbors $k$ to 10 for all experiments (the paper shows that the performance of kNN-Per is robust to this value). As mentioned in Section~\ref{sec:cross-device-exp-knn}, we tune the interpolation coefficient globally for all clients (i.e., every client use the same coefficient), because tuning per-client specific coefficient performed worse. The tuned interpolation hyperparameters are shown below (best values are highlighted in \verb|**value**|):

\paragraph{EMNIST}\mbox{}\\

\verb|coefficient=[0, 0.1, 0.2, 0.3, **0.4**, 0.5, 0.6, 0.7, 0.8, 0.9, 1.0]|

\paragraph{StackOverflow}\mbox{}\\

\verb|coefficient=[0, 0.1, **0.2**, 0.3, 0.4, 0.5, 0.6, 0.7, 0.8, 0.9, 1.0]|

\paragraph{Landmarks}\mbox{}\\

\verb|coefficient=[0, 0.1, 0.2, 0.3, 0.4, **0.5**, 0.6, 0.7, 0.8, 0.9, 1.0]|

\paragraph{TedMulti-EnEs}\mbox{}\\

\verb|coefficient=[0, **0.1**, 0.2, 0.3, 0.4, 0.5, 0.6, 0.7, 0.8, 0.9, 1.0]|

\subsubsection{Local training}

Traning a local model at each client can be done by running
\verb|finetuning_trainer.py| with \verb|total_rounds=0|. Note that what happens is that every client fine-tunes a random model (sent by the server) locally. As long as
we set a large enough \verb|finetune_max_epochs| (the best number of local epochs will be found after postprocessing the validation metrics), this will give the desired
metrics where each client learns a local model without federation.

\paragraph{EMNIST}\mbox{}\\

Fixed hyperparameters:
\begin{itemize}
    \item \verb|total_rounds=[0]|
    \item \verb|valid_clients_per_round=[100]|
    \item \verb|test_clients_per_round=[100]|
    \item \verb|finetune_optimzier=[‘sgd’]|
    \item \verb|finetune_max_epochs=[200]|
\end{itemize}

Tuned hyperparameters (best values are highlighted in \verb|**value**|):
\begin{itemize}
    \item \verb|finetune_last_layer=[True, **False**]|
    \item \verb|finetune_learning_rate=[0.001, 0.01, **0.1**, 0.5, 1.0]|
\end{itemize}

\paragraph{StackOverflow}\mbox{}\\

Fixed hyperparameters:
\begin{itemize}
    \item \verb|total_rounds=[0]|
    \item \verb|valid_clients_per_round=[200]|
    \item \verb|test_clients_per_round=[200]|
    \item \verb|finetune_optimzier=[‘sgd’]|
    \item \verb|finetune_max_epochs=[200]|
\end{itemize}

Tuned hyperparameters (best values are highlighted in \verb|**value**|):
\begin{itemize}
    \item \verb|finetune_last_layer=[True, **False**]|
    \item \verb|finetune_learning_rate=[0.1, **0.5**, 1.0]|
\end{itemize}

\paragraph{Landmarks}\mbox{}\\

Fixed hyperparameters:
\begin{itemize}
    \item \verb|total_rounds=[0]|
    \item \verb|valid_clients_per_round=[32]|
    \item \verb|test_clients_per_round=[96]|
    \item \verb|finetune_optimzier=[‘sgd’]|
    \item \verb|finetune_max_epochs=[50]|
\end{itemize}

Tuned hyperparameters (best values are highlighted in \verb|**value**|):
\begin{itemize}
    \item \verb|finetune_last_layer=[True, **False**]|
    \item \verb|finetune_learning_rate=[0.0001, 0.001, **0.01**, 0.1]|
\end{itemize}

\paragraph{TedMulti-EnEs}\mbox{}\\

Fixed hyperparameters:
\begin{itemize}
    \item \verb|total_rounds=[0]|
    \item \verb|valid_clients_per_round=[98]|
    \item \verb|test_clients_per_round=[117]|
    \item \verb|finetune_optimzier=[‘sgd’]|
    \item \verb|finetune_max_epochs=[50]|
\end{itemize}

Tuned hyperparameters (best values are highlighted in \verb|**value**|):
\begin{itemize}
    \item \verb|finetune_last_layer=[**True**, False]|
    \item \verb|finetune_learning_rate=[0.5, **1.0**, 2.0, 3.0]|
\end{itemize}

\subsection{Cross-silo} 
\label{app:hparam:silo}

The definitions of hyperparameters can be found in \verb|main.py| in the implementation folders for the Vehicle, School, and ADNI datasets respectively at \url{https://github.com/google-research/federated/tree/master/personalization_benchmark/cross_silo}.

\textbf{Common hyperparameters.} 
The following hyperparameters are fixed across all cross-silo experiments:
\begin{itemize}
    \item \verb|client_optimizer=[`sgd']|
    \item \verb|server_optimizer=[`fedavgm']|
    \item \verb|fedavgm_momentum=[0.9]|
    \item \verb|inner_epochs=[1]| \\
    (This is the number of local training epochs performed by a client during a round of training. For methods that train a global model, this is the number of local training epochs before returning the model update to the server.)
\end{itemize}
We focus on FedAvgM~\cite{hsu2019measuring} for simplicity of hyperparameter tuning. Note also that the precise variable names of the hyperparameters (typefaced with \verb|monospaced font|) may vary depending on the dataset-specific implementation.

\textbf{Computational resources.}
For the Vehicle and School datasets, all experiments (including hyperparameter search) are done on 88 commodity CPU cores. 
For the ADNI dataset, each run is done on one GPU.

\subsubsection{Local training}

\paragraph{Vehicle}\mbox{}\\

Fixed hyperparameters:
\begin{itemize}
    \item \verb|num_rounds=[500]|
    \item \verb|clients_per_round=[23]|
    \item \verb|batch_size=[64]|  (client local training batch size)
\end{itemize}

Tuned hyperparameters (best values are highlighted in \verb|**value**|):

\begin{itemize}
    \item \verb|client_lrs=[0.003, 0.01, **0.03**, 0.1, 0.3]|
\end{itemize}

\paragraph{School}\mbox{}\\

Fixed hyperparameters:

\begin{itemize}
    \item \verb|num_rounds=[500]|
    \item \verb|clients_per_round=[139]|
    \item \verb|batch_size=[32]|  (client local training batch size)
\end{itemize}

Tuned hyperparameters (best values are highlighted in \verb|**value**|):

\begin{itemize}
    \item \verb|client_lrs=[0.001, 0.003, **0.01**, 0.03, 0.1]|
\end{itemize}

\paragraph{ADNI}\mbox{}\\

Fixed hyperparameters:

\begin{itemize}
    \item \verb|num_rounds=[70]|
    \item \verb|clients_per_round=[9]|
    \item \verb|batch_size=[64]|  (client local training batch size)
\end{itemize}

Tuned hyperparameters (best values are highlighted in \verb|**value**|):

\begin{itemize}
    \item \verb|client_lrs=[0.001, **0.01**, 0.1]|
\end{itemize}

\subsubsection{FedAvg+Fine-tuning}

\paragraph{Vehicle}\mbox{}\\

Fixed hyperparameters:

\begin{itemize}
    \item \verb|num_rounds=[500]|
    \item \verb|finetune_epochs=[100]|  (max number of local epochs for fine-tuning)
    \item \verb|finetune_every=[50]|  (run fine-tuning every number of rounds)
    \item \verb|finetune_optimizer=[`sgd']|
    \item \verb|clients_per_round=[23]|
    \item \verb|batch_size=[64]|  (client local training batch size)
\end{itemize}

Tuned hyperparameters (best values are highlighted in \verb|**value**|):

\begin{itemize}
    \item \verb|client_lrs=[**0.003**, 0.01, 0.03, 0.1, 0.3]|
    \item \verb|server_lrs=[0.1, 0.3, 1, 3, **10**]|
    \item \verb|finetune_lrs=[0.003, 0.01, 0.03, 0.1, 0.3]|  \\
    (The fine-tuning client learning rates are tuned separately for each client.)
\end{itemize}

\paragraph{School}\mbox{}\\

Fixed hyperparameters:

\begin{itemize}
    \item \verb|num_rounds=[500]|
    \item \verb|finetune_epochs=[100]|  (max number of local epochs for fine-tuning)
    \item \verb|finetune_every=[50]|  (run fine-tuning every number of rounds)
    \item \verb|finetune_optimizer=[`sgd']|
    \item \verb|clients_per_round=[139]|
    \item \verb|batch_size=[32]|  (client local training batch size)
\end{itemize}

Tuned hyperparameters (best values are highlighted in \verb|**value**|):

\begin{itemize}
    \item \verb|client_lrs=[**0.001**, 0.003, 0.01, 0.03, 0.1]|
    \item \verb|server_lrs=[0.1, 0.3, 1, **3**, 10]|
    \item \verb|finetune_lrs=[0.001, 0.003, 0.01, 0.03, 0.1]| \\
    (The fine-tuning client learning rates are tuned separately for each client.)
\end{itemize}

\paragraph{ADNI}\mbox{}\\

Fixed hyperparameters:
\begin{itemize}
    \item \verb|num_rounds=[70]|
    \item \verb|finetune_optimzier=[`sgd']|
    \item \verb|clients_per_round=[9]|
    \item \verb|batch_size=[64]|  (client local training batch size)
\end{itemize}

Tuned hyperparameters (best values are highlighted in \verb|**value**|):
\begin{itemize}
    \item \verb|finetune_last_layer=[True, False]|
    \item \verb|server_lrs=[1, 3, **10**, 20]|
    \item \verb|client_lrs=[0.0001, 0.001, **0.01**, 0.1]|
    \item \verb|finetune_lrs=[0.0001, 0.001, 0.01, 0.1]| \\
    (When running FedAvg, the optimal client-side learning rate is 0.01. During fine-tuning, we tune client learning rates (i.e., fine-tuning learning rates) separately for each client.)
\end{itemize}

\subsubsection{HypCluster}

\paragraph{Vehicle}\mbox{}\\

Fixed hyperparameters:

\begin{itemize}
    \item \verb|num_rounds=[500]|
    \item \verb|clients_per_round=[23]|
    \item \verb|batch_size=[64]|  (client local training batch size)
\end{itemize}

Tuned hyperparameters (best values are highlighted in \verb|**value**|):

\begin{itemize}
    \item \verb|client_lrs=[0.003, **0.01**, 0.03, 0.1, 0.3]|
    \item \verb|server_lrs=[0.1, 0.3, 1, 3, **10**]|
    \item \verb|num_clusters=[2, 3, **4**]|
    \item \verb|warmstart_fracs=[**0**, 0.2]| \\
    (The fraction of rounds for warm starting; the rest runs clustering training.)
\end{itemize}

\paragraph{School}\mbox{}\\

Fixed hyperparameters:

\begin{itemize}
    \item \verb|num_rounds=[500]|
    \item \verb|clients_per_round=[139]|
    \item \verb|batch_size=[32]|  (client local training batch size)
\end{itemize}

Tuned hyperparameters (best values are highlighted in \verb|**value**|):

\begin{itemize}
    \item \verb|client_lrs=[0.001, 0.003, **0.01**, 0.03, 0.1]|
    \item \verb|server_lrs=[0.1, 0.3, 1, 3, **10**]|
    \item \verb|num_clusters=[2, **3**, 4]|
    \item \verb|warmstart_fracs=[0, **0.2**]| \\
    (The fraction of rounds for warm starting; the rest runs clustering training.)
\end{itemize}

\paragraph{ADNI}\mbox{}\\

Fixed hyperparameters:

\begin{itemize}
    \item \verb|num_rounds=[70]|
    \item \verb|warm_start_rounds=[20]|
    \item \verb|clustering_training_rounds=[50]|
    \item \verb|clients_per_round=[9]|
    \item \verb|batch_size=[64]|  (client local training batch size)
\end{itemize}

Tuned hyperparameters (best values are highlighted in \verb|**value**|):

\begin{itemize}
    \item \verb|server_lrs=[1, 3, **10**, 20]|
    \item \verb|client_lrs=[0.0001, 0.001, **0.01**, 0.1]|
    \item \verb|num_clusters=[**2**, 3, 4]|
\end{itemize}

\subsection{FedAvg + kNN-Per}

We use the model trained by FedAvg (same as "FedAvg + Fine-tuning") as the global model for personalization. Following~\cite{marfoq2022personalized}, we set the number of nearest neighbors $k$ to 10 for all experiments (the paper shows that the performance of kNN-Per is robust to this value). Similar to the cross-device setting, we also tune the interpolation coefficient globally for all clients (i.e., every client use the same coefficient). The tuned interpolation coefficients are shown below (best values are highlighted in \verb|**value**|):

\paragraph{Vehicle}\mbox{}\\

\verb|coefficient=[0, 0.1, 0.3, 0.5, 0.7, **0.9**, 1.0]|

\paragraph{School}\mbox{}\\

\verb|coefficient=[0, 0.1, 0.3, **0.5**, 0.7, 0.9, 1.0]|

\paragraph{ADNI}\mbox{}\\

\verb|coefficient=[0, 0.1, 0.3, **0.5**, 0.7, 0.9, 1.0]|

\subsubsection{MOCHA}

Note that MOCHA~\cite{smith2017federated} was implemented in its primal form (i.e.\ gradient descent update for the local personalized models, with task-relationship learning regularization) instead of the dual form since the dual was not derived for regression in the original paper. We were able to reproduce the results reported in the original paper in the primal (with an average error rate of 6.29 in \Cref{table:cross-silo,apptable:cross-silo}, smaller than 6.59 that was reported). In the primal, MOCHA has optimization hyperparameters such as client optimizer, client learning rates, batch sizes, etc. as with other methods.

\paragraph{Vehicle}\mbox{}\\

Fixed hyperparameters:

\begin{itemize}
    \item \verb|num_rounds=[500]|
    \item \verb|clients_per_round=[23]|
    \item \verb|batch_size=[64]|  (client local training batch size)
\end{itemize}

Tuned hyperparameters (best values are highlighted in \verb|**value**|; an alternative combination that gives very similar results are highlighted in \verb|_values_|):

\begin{itemize}
    \item \verb|client_lrs=[0.003, _0.01_, **0.03**, 0.1, 0.3]|
    \item \verb|lambdas=[**0.0001**, _0.001_, 0.01, 0.1, 0.3, 1, 3]| \\
    (MTL regularization strength.)
    \item \verb|mocha_outers=[**1**, _2_, 5]| \\
    (The number of local epochs every server update of the task-relationship matrix. This nests with \verb|inner_epochs| and may thus increase the total number of epochs over the local datasets.)
\end{itemize}

\paragraph{School}\mbox{}\\

Fixed hyperparameters:

\begin{itemize}
    \item \verb|num_rounds=[500]|
    \item \verb|clients_per_round=[139]|
    \item \verb|batch_size=[32]|  (client local training batch size)
\end{itemize}

Tuned hyperparameters (best values are highlighted in \verb|**value**|; an alternative combination that gives very similar results are highlighted in \verb|_values_|):

\begin{itemize}
    \item \verb|client_lrs=[0.001, **0.003**, _0.01_, 0.03, 0.1]|
    \item \verb|lambdas=[**0.0001**, 0.001, _0.01_, 0.1, 0.3, 1, 3]| \\
    (MTL regularization strength.)
    \item \verb|mocha_outers=[_1_, **2**, 5]| \\
    (See Vehicle hyperparameters above for description.)
\end{itemize}

\subsubsection{Ditto}

\paragraph{Vehicle}\mbox{}\\

Fixed hyperparameters:

\begin{itemize}
    \item \verb|num_rounds=[500]|
    \item \verb|personalized_model_inner_epochs=[1]|
    \item \verb|clients_per_round=[23]|
    \item \verb|batch_size=[64]|  (client local training batch size)
\end{itemize}

Tuned hyperparameters (best values are highlighted in \verb|**value**|):

\begin{itemize}
    \item \verb|client_lrs=[0.003, 0.01, 0.03, 0.1, **0.3**]|
    \item \verb|server_lrs=[0.1, 0.3, **1**, 3, 10]|
    \item \verb|lambdas=[0.0001, 0.001, **0.01**, 0.1, 0.3, 1, 3]| \\
    (MTL regularization strength.)
    \item \verb|personalized_model_lrs=[0.003, 0.01, 0.03, **0.1**, 0.3]| \\
    (The client learning rate for Ditto's personalized models. This may be different from the client learning rates used to update the global model.)
\end{itemize}

\paragraph{School}\mbox{}\\

Fixed hyperparameters:

\begin{itemize}
    \item \verb|num_rounds=[500]|
    \item \verb|personalized_model_inner_epochs=[1]|
    \item \verb|clients_per_round=[139]|
    \item \verb|batch_size=[32]|  (client local training batch size)
\end{itemize}

Tuned hyperparameters (best values are highlighted in \verb|**value**|):

\begin{itemize}
    \item \verb|client_lrs=[**0.001**, 0.003, 0.01, 0.03, 0.1]|
    \item \verb|server_lrs=[0.1, 0.3, 1, 3, **10**]|
    \item \verb|lambdas=[0.0001, 0.001, 0.01, 0.1, 0.3, **1**, 3]| \\
    (MTL regularization strength.)
    \item \verb|personalized_model_lrs=[0.001, 0.003, **0.01**, 0.03, 0.1]| \\
    (See Vehicle hyperparameters above for description.)
\end{itemize}

\paragraph{ADNI}\mbox{}\\

Fixed hyperparameters:

\begin{itemize}
    \item \verb|num_rounds=[100]|
    \item \verb|clients_per_round=[9]|
    \item \verb|local_batch_size=[64]|
    \item \verb|personalized_model_local_epoch=[1]|
\end{itemize}

Tuned hyperparameters (best values are highlighted in \verb|**value**|):

\begin{itemize}
    \item \verb|server_lrs=[1, 3, **10**, 20]|
    \item \verb|client_lrs=[0.0001, 0.001, **0.01**, 0.1]|
    \item \verb|personalized_model_lrs=[0.0001, **0.001**, 0.01]| \\
    (See Vehicle hyperparameters above for description.)
    \item \verb|lambdas=[0.01, 0.1, **1**]| \\ 
    (MTL regularization strength.)
\end{itemize}

\section{Additional Experimental Results and Discussions}\label{app:more-exp}

In this section, we provide a few more experimental results that are omitted from the main paper due to space limitation.

\textbf{Cross-device experiments.}
Table~\ref{apptable:cross-device} extends Table~\ref{table:cross-device} with standard deviations across 5 different runs for each metric. 
Note that there are two standard deviations here:
1) the standard deviation of per-client accuracy across the test clients, which is used as a fairness metric\footnoteref{fairness} and is already included in Table~\ref{table:cross-device};
2) the standard deviation of each metric value (e.g., the standard deviation of the fairness metric mentioned in 1)) across 5 different runs (each run uses a different seed), which is omitted from the main paper and is included in Table~\ref{apptable:cross-device}.

\begin{table}[t!]
  \renewcommand{\arraystretch}{1.1}
  \caption{Complete results of the experiments in Table~\ref{table:cross-device}. The only difference is that we report each metric's mean and standard deviation (std) across 5 different runs. Note that there are two stds here: 1) "per-client acc std", the std of per-client accuracy across the test clients, which is a fairness metric considered in~\cite{li2021ditto, mansour2020three} and is already included in Table~\ref{table:cross-device}; 2) the std across 5 different runs, which is omitted from Table~\ref{table:cross-device} and is shown after $\pm$ in this table.}
  \label{apptable:cross-device}
  \centering
  \scalebox{1}{
  \begin{tabular}{p{1.8cm}|p{4.5cm}p{1.9cm}p{2.2cm}p{2cm}p{1.6cm}} \toprule
    \textbf{Algorithm} & \textbf{Metrics} & \textbf{EMNIST} & \textbf{StackOverflow} & \textbf{Landmarks} & \textbf{TedMulti}\\
    \midrule
    \multirow{2}{1.8cm}{Local} & Per-client acc mean & 0.594$\pm$.011 &0.062$\pm$.001&0.173$\pm$.008&0.056$\pm$.0004 \\
    & Per-client acc std & 0.17$\pm$.011 &0.02$\pm$.005&0.16$\pm$.008&0.02$\pm$.0003 \\
    \midrule
    \multirow{8}{1.8cm}{FedAvg + Fine-tuning (FT)}&Per-client acc mean before FT & 0.844$\pm$.004 & 0.269$\pm$.002 & 0.564$\pm$.008 &0.160$\pm$.002\\
    &Per-client acc std before FT & 0.1$\pm$.004 & 0.03$\pm$.0007 & 0.16$\pm$.005 &0.04$\pm$.004\\
    &Per-client acc after FT & 0.903$\pm$.002 & 0.282$\pm$.002& 0.773$\pm$.006 &0.162$\pm$.002 \\
    &Per-client acc std after FT & 0.06$\pm$.0005 & 0.03$\pm$.001 & 0.11$\pm$.006 &0.04$\pm$.001\\
    &\% clients "hurt" after FT &5.2\%$\pm$0.7\%&14\%$\pm$2.4\%&5.6\%$\pm$2.4\%&40\%$\pm$4.8\%\\
    &FT all layers vs last layer & All layers & All layers & All layers &All layers\\ 
    \cmidrule{2-6}
    &\multicolumn{5}{c}{\textit{\textbf{Practical concerns}: difficult to tune hyperparameters; may hurt clients; sensitive}}\\
    &\multicolumn{5}{c}{\textit{to distribution shift; performance drops with fewer local examples (see Section~\ref{sec:cross-device-exp-FT})}}\\
    \midrule
    \multirow{10}{1.8cm}{HypCluster / IFCA}&Per-client acc mean & 0.897$\pm$.002 &0.273$\pm$.0009& 0.573$\pm$.005 &0.163$\pm$.004\\
    &Per-client acc std & 0.08$\pm$.003 &0.03$\pm$.002& 0.16$\pm$.003 &0.04$\pm$.002\\
    & No. tuned clusters ($k$)& 2 & 2 & 2 & 2\\
    & \% clients largest cluster & 52.6\%$\pm$0.8\% &85.1\%$\pm$1.6\%&92.1\%$\pm$3.6\%&54.7\%$\pm$0\\
    & Warmstart from FedAvg & Yes & Yes & Yes&Yes\\
    & Per-client acc mean by ensembling $k$ FedAvg models & 0.860$\pm$.005&0.271$\pm$.002&0.564$\pm$.016&0.163$\pm$.004\\
    & Per-client acc std by ensembling $k$ FedAvg models & 0.08$\pm$.006&0.03$\pm$.002&0.16$\pm$.007&0.04$\pm$.004\\\cmidrule{2-6}
    &\multicolumn{5}{c}{\textit{\textbf{Practical concerns}: difficult to train due to mode collapse; high communication cost;}}\\
    &\multicolumn{5}{c}{\textit{difficult to interpret the learned clusters; sensitive to distribution shift (see Section~\ref{sec:cross-device-exp-cluster})}}\\
    \midrule
    \multirow{5}{1.8cm}{FedAvg + kNN-Per}&Per-client acc mean& 0.876$\pm$.003 & 0.275$\pm$.0005 & 0.735$\pm$.007 & 0.162$\pm$.002\\
    & Per-client acc std& 0.06$\pm$.004 & 0.03$\pm$.0009 & 0.13$\pm$.005 & 0.05$\pm$.004 \\
    &\% clients "hurt"
    &19.8\%$\pm$4.2\% & 23.6\%$\pm$ 3.4\% &5.6\%$\pm$2.3\% &34.4\%$\pm$4.5\% \\
    \cmidrule{2-6}
    &\multicolumn{5}{c}{\textit{\textbf{Practical concerns}: difficult to tune hyperparameters; may hurt clients; sensitive}}\\
    &\multicolumn{5}{c}{\textit{to distribution shift; performance drops with fewer local examples (see Section~\ref{sec:cross-device-exp-knn})}}\\
    \bottomrule 
    \end{tabular}}
\end{table}

\begin{table}[t!]
  \renewcommand{\arraystretch}{1.1}
  \caption{Complete results of the experiments in Table~\ref{table:cross-silo}. The only difference is that we report each metric's mean and standard deviation (std) across 5 different runs. Similar to results in cross-device experiments, there are two stds here: 1) "per-client acc std", the std of per-client accuracy across the test clients, which is a fairness metric considered in~\cite{li2021ditto, mansour2020three} and is already included in Table~\ref{table:cross-silo}; 2) the std across 5 different runs, which is omitted from Table~\ref{table:cross-silo} and is shown after $\pm$ in this table. 
  } 
  \label{apptable:cross-silo}
  \centering
  \scalebox{1}{
  \begin{tabular}{p{2cm}|p{5.2cm}p{2.4cm}p{2.4cm}p{2.4cm}}
    \toprule
    \textbf{Algorithm} & \textbf{Metrics}  & \textbf{Vehicle} (\textbf{\textcolor{OliveGreen}{acc}}) & \textbf{School} (\textbf{\textcolor{BrickRed}{mse}}) & \textbf{ADNI} (\textbf{\textcolor{BrickRed}{mse}}) \\
    \midrule
    \multirow{2}{1.8cm}{Local training} & Per-client metric mean & 0.9367$\pm$.0248  & 0.0121$\pm$.0059 & 0.0177$\pm$.0007 \\
     & Per-client metric std & 0.0027$\pm$.0029  &  0.0003$\pm$.0008  & 0.0106$\pm$.0007 \\
    \midrule
    \multirow{8}{1.8cm}{FedAvg + Fine-tuning (FT)} & Per-client metric mean before FT & 0.8859$\pm$.0833  & 0.0130$\pm$.0068 & 0.0141$\pm$.0004  \\
     & Per-client metric std before FT & 0.0033$\pm$.0028  & 0.0002$\pm$.0009 & 0.0091$\pm$.0012  \\
    & Per-client metric mean after FT  &  0.9385$\pm$.0253  & 0.0116$\pm$.0056 & 0.0124$\pm$.0007 \\
    & Per-client metric std after FT  &  0.0029$\pm$.0015  & 0.0002$\pm$.0006 & 0.0082$\pm$.0012 \\
    & \% clients "hurt" after FT  & 4.4\%$\pm$3.9\% &  33.0\%$\pm$3.0\% & 0$\pm$0  \\
    & FT all layers vs last layer & N/A & N/A  & no difference \\
    \cline{2-5}
    &\multicolumn{4}{c}{\textit{\textbf{Practical concerns:} Similar to those in the cross-device setting (Table~\ref{table:cross-device});}} \\
    &\multicolumn{4}{c}{\textit{Tuning per-client FT hyperparameter may outperform tuning globally (\Cref{fig:cross-silo}).}} \\
    \midrule
    \multirow{9}{1.8cm}{HypCluster / IFCA} & Per-client metric mean & 0.9246$\pm$.0288  & 0.0112$\pm$.0053  &  0.0137$\pm$.0012  \\
    & Per-client metric std & 0.0058$\pm$.0043  & 0.0003$\pm$.0006  &  0.0093$\pm$.0017  \\
    & No. tuned clusters ($k$) & 4 & 3  & 2 \\
    & \% clients largest cluster  & 49.6\%$\pm$3.5\% & 44.6\%$\pm$1.9\% &  78\%$\pm$17.6\%  \\
    & Warmstart from FedAvg & No & Yes & Yes \\
    & Per-client metric mean by ensembling $k$ FedAvg models  &  0.8851$\pm$.0828  &  0.0129$\pm$.0066  &  0.0133$\pm$.0006 \\
    & Per-client metric std by ensembling $k$ FedAvg models  &  0.0028$\pm$.0032  &  0.0001$\pm$.0008  &  0.0091$\pm$.0011 \\
    \cline{2-5}
    &\multicolumn{4}{c}{\textit{\textbf{Practical concerns:} Similar to those in the cross-device setting (Table~\ref{table:cross-device}); on Vehicle}}\\
    &\multicolumn{4}{c}{\textit{and School, mode collapse may occur less potentially because of linear model.} }\\
    \midrule
    \multirow{4}{2cm}{FedAvg+kNN-Per} & Per-client metric mean & 0.9228$\pm$.0028  & 0.01163$\pm$.0002 & 0.0126$\pm$.0008  \\
    & Per-client metric std & 0.0287$\pm$.0016 & 0.0055$\pm$.0006 & 0.0096$\pm$.0006 \\
    & \% clients "hurt"  &  22.6\%$\pm$4.3\%  & 37.4\%$\pm$0.6\% & 37.8\%$\pm$16.8\% \\
    \cline{2-5}
    &\multicolumn{4}{c}{\textit{\textbf{Practical concerns:} Similar to those in the cross-device setting (Table~\ref{table:cross-device}).}} \\
    \midrule
    \multirow{2}{2cm}{MTL (Ditto)} & Per-client metric mean & 0.9377$\pm$.0218 & 0.0114$\pm$.0053  &  0.0134$\pm$.0004   \\
     & Per-client metric std & 0.0025$\pm$.0026 & 0.0002$\pm$.0006  & 0.0074$\pm$.0008   \\
     \multirow{2}{2.2cm}{MTL (Mocha)} & Per-client metric  mean &  0.9371$\pm$.0244 & 0.0121$\pm$.0059  & N/A    \\
     & Per-client metric std &  0.0030$\pm$.0025 & 0.0003$\pm$.0009  & N/A    \\
    \bottomrule
  \end{tabular}}
\end{table}

\begin{figure}[ht]
    \centering
    \includegraphics[width=0.9\linewidth]{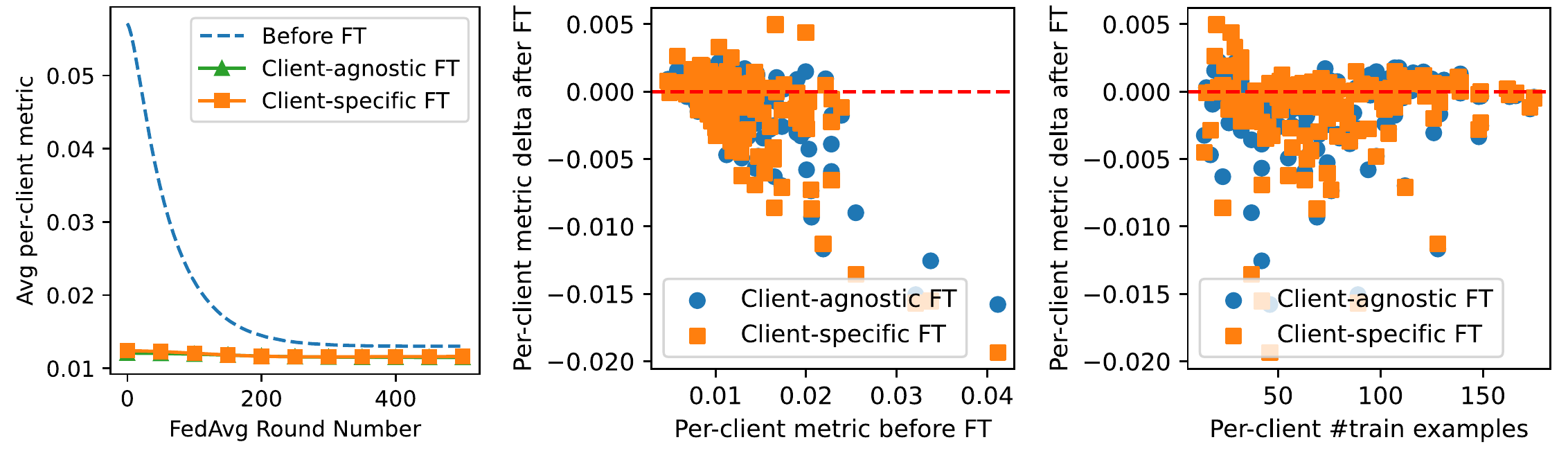}
    \caption{
    Effects of client-specific fine-tuning on the School dataset (each client chooses a custom fine-tuning learning rate and stopping epoch). Note that School uses MSE (mean squared error) as the evaluation metric here, thus the \textbf{more negative} the per-client metric delta, the better. Unlike Figure~\ref{fig:cross-silo}, tuning the fine-tuning hyperparameters globally and tuning in a per-client manner performs the same on the School dataset. This is potentially due to the fact that the client's local dataset is relatively small for School (as shown in Table~\ref{table:benchmark-overview}).
    }
    \label{appfig:school-client-specific-ft}
\end{figure}

\textbf{Cross-silo experiments.}
Similar to the cross-device counterpart, Table~\ref{apptable:cross-silo} extends Table~\ref{table:cross-silo} with standard deviation across 5 different runs for each metric.
One small caveat around the use of client-specific fine-tuning hyperparameters, as discussed in \Cref{sec:cross-silo-exp} of the main paper, is that it may improve performance only when the clients' local datasets are large. 
Here, we perform an additional analysis on the School dataset, where the local datasets are relatively small (recall Table~\ref{table:benchmark-overview}).
\Cref{appfig:school-client-specific-ft} visualizes the statistics of clients hurt when training on School, akin to \Cref{fig:clients-hurt} (note that we use MSE here, so a positive delta means this client gets hurt by fine-tuning).
Observe that while many clients benefited from the client-specific fine-tuning (those with negative metric deltas), some clients do not observe an improvement and may even see a slight degradation in their utility compared to the client-agnostic fine-tuning setting. We argue that this is due to clients having small local datasets such that the selected client-specific FT hyperparameters may overfit to the small local validation sets and may not reflect general improvement on the local test sets.
Note also that while Table~\ref{table:cross-silo} and Table~\ref{apptable:cross-silo} suggests a seemingly considerable fraction (33\%) of clients is hurt after fine-tuning, we can observe from \Cref{appfig:school-client-specific-ft} that most `hurt' clients have borderline metric changes.
Overall, these observations suggest that the effect of client-specific fine-tuning can be mixed if the client local datasets are relatively small, as is usually the case with cross-device federated learning.

\clearpage

\section{Conclusion}\label{app:conclusion}
We present \name, a large-scale benchmark of personalized federated learning covering both cross-device and cross-silo settings. \name provides an end-to-end experimental pipeline including data preprocessing, algorithms, evaluation metrics, and tuned hyperparameters\footnoteref{code}. Beyond these baselines, our experiments provide new insights about personalized FL and suggest several directions of future work.
\begin{itemize}[leftmargin=*]
    \item The notion of the "best" method (or even the "best" hyperparameter of the same method) can change when we change the evaluation metrics or settings. In the cross-device setting, FedAvg+Fine-tuning achieves the best average per-client accuracy for 3/4 datasets, and at the same time also improves fairness\footnoteref{fairness} (Table~\ref{table:cross-device} and Figure~\ref{fig:box-plot}). The best hyperparameters of FedAvg+Fine-tuning can change if we look at different metrics (Figure~\ref{fig:tune_ft_hparams}). On the TedMulti-EnEs dataset, it is difficult to determine which personalization algorithm is the "best": while HypCluster achieves the best average per-client accuracy, it is worse than FedAvg+Fine-tuning from the perspective of accuracy-communication ratio (Figure~\ref{fig:hypcluster}(c)); on the other hand, the fine-tuned models of $40\%$ clients on TedMulti-EnEs are worse than that of FedAvg (Table~\ref{table:cross-device}). In the cross-silo setting, both FedAvg+Fine-tuning and MTL achieve the best average per-client accuracy over the Vehicle and School datasets (Table~\ref{table:cross-silo}). MTL may have an additional advantage of having less hyperparameters than FedAvg+Fine-tuning. On the Vehicle dataset, local training seems the "best" because it achieves a similar accuracy as other methods but is more private. On the ADNI dataset, both FedAvg+Fine-tuning and FedAvg+kNN-per achieve the best per-client MSE, but kNN-Per hurts more clients than fine-tuning. Given that the notion of "best" method can change depending on the metrics or settings, a critical future direction is thus to develop systematic evaluation schemes for personalized FL (i.e., mean accuracy alone is not enough).
    \item Existing literature often overlook or obfuscate the practical complexities of deploying personalized FL algorithms in real-world settings. For example, in Section~\ref{sec:cross-device-exp-FT}, we discuss that local data scarcity and heterogeneity create a fundamental challenge to tuning FedAvg+Fine-tuning hyperparameters in the cross-device setting. Because each client has a very small local dataset in the cross-device setting, the fine-tuning hyperparameters are tuned globally (instead of in a per-client manner as in the cross-silo setting). The globally tuned hyperparameters may hurt some clients (Figure~\ref{fig:clients-hurt}). Ideally, we want to find a good set of fine-tuning hyperparameters such that the overall improvement is large and no clients are hurt after fine-tuning, which can be fundamentally difficult (Figure~\ref{fig:tune_ft_hparams}). Given the observed benefits of per-client hyperparameter tuning in cross-silo FL (Figure~\ref{fig:cross-silo}), it may be beneficial to develop similar, scalable approaches for hyperparameter tuning in cross-device FL. Taking these practical considerations into account when designing new personalized FL algorithms is crucial in making these algorithms more effective in practice.
    \item Improving HypCluster requires solving the mode collapse issue and rethinking the difference between clustering and ensembling. How to effectively train HypCluster in the real-world federated learning systems is an interesting open problem (Figure~\ref{fig:hypcluster}(a)). Although warmstart can mitigate the mode collapse issue, Table~\ref{table:cross-device} and Table~\ref{table:cross-silo} show that the performance of HypCluster is similar to that of ensembling\footnoteref{ensemble} multiple models learned by FedAvg. Does HypCluster really capture the underlying clustering structure (Figure~\ref{fig:hypcluster}(d))? Answering this question requires interpreting the learned clusters in the federated learning setting.  
    \item Tradeoffs exists between adapting a client’s personalized model to the current local distribution and generalizing to future distributions. As shown in Figure~\ref{fig:id_ood_tradeoff}, this tradeoff between personalization and generalization exists for all personalized FL algorithms, which is worth exploring in greater detail (see also the remark on "acceptance criteria and robust personalization" in Section~\ref{sec:cross-device-exp-FT}). 
\end{itemize}


\end{document}